\documentclass[conference]{IEEEtran}
\IEEEoverridecommandlockouts
\usepackage{cite}
\usepackage{amsmath,amssymb,amsfonts}
\usepackage{algorithm}
\usepackage{algpseudocode}
\usepackage{graphicx}
\usepackage{subfigure}
\usepackage{textcomp}
\usepackage{xcolor}
\usepackage{bm}
\usepackage{epstopdf}

\def\BibTeX{{\rm B\kern-.05em{\sc i\kern-.025em b}\kern-.08em
    T\kern-.1667em\lower.7ex\hbox{E}\kern-.125emX}}
    
\begin{document}

\title{SAFE: Scalable Automatic Feature Engineering Framework for Industrial Tasks}

\author{
\IEEEauthorblockN{Qitao Shi, Ya-Lin Zhang$\dagger$, Longfei Li, Xinxing Yang, Meng Li, Jun Zhou \thanks{$\dagger$: corresponding author.}}
\IEEEauthorblockA{
	Ant Financial Services Group \\
	\{qitao.sqt, lyn.zyl, longyao.llf, xinxing.yangxx, lm168260, jun.zhoujun\}@antfin.com}
}

\maketitle

\begin{abstract}

Machine learning techniques have been widely applied in Internet companies for various tasks, acting as an essential driving force, and feature engineering has been generally recognized as a crucial tache when constructing machine learning systems. Recently, a growing effort has been made to the development of automatic feature engineering methods, so that the substantial and tedious manual effort can be liberated. However, for industrial tasks, the efficiency and scalability of these methods are still far from satisfactory. In this paper, we proposed a staged method named SAFE (\textbf{S}calable \textbf{A}utomatic \textbf{F}eature \textbf{E}ngineering), which can provide excellent efficiency and scalability, along with requisite interpretability and promising performance. Extensive experiments are conducted and the results show that the proposed method can provide prominent efficiency and competitive effectiveness when comparing with other methods. What's more, the adequate scalability of the proposed method ensures it to be deployed in large scale industrial tasks.

\end{abstract}


\bstctlcite{IEEEexample:BSTcontrol}
\bstctlcite{MyBSTcontrol} 

\section{Introduction}
\label{Introduction}

Nowadays, machine learning (ML) techniques have been widely explored and applied in almost all Internet companies, and serving as essential parts in diversified fields, such as recommendation system~\cite{DBLP:conf/recsys/DavidsonLLNVGGHLLS10,DBLP:conf/adaptive/PazzaniB07,DBLP:conf/kdd/HePJXLXSAHBC14}, fraud detection~\cite{DBLP:journals/csur/ChandolaBK09,DBLP:conf/www/ZhangLZLZ18,DBLP:journals/tist/ZhangZZFLLLZCLQ19}, advertising~\cite{DBLP:conf/sigir/LacerdaCGFZR06,DBLP:conf/sigir/Ribeiro-NetoCGM05,DBLP:conf/kdd/ZhuJTPZLG17}, and face recognition~\cite{DBLP:journals/access/AlajmiAEAF19,DBLP:conf/ijcnn/YeungLC17}, etc. 
With the help of these techniques, excellent performance and significant improvement have been obtained.

Generally speaking, to build a machine learning system, a professional and complex ML pipeline is always needed, which usually includes data preparation, feature engineering, model generation, and model evaluation, etc.
It is widely agreed that the performance of machine learning methods depends to a large extent on the quality of the features, and generating a good feature set becomes a crucial step to chase high performance~\cite{DBLP:books/lib/HastieTF09}. 
Therefore, most machine learning engineers take a large effort to obtain useful features when building a machine learning system. 

However, it is frustrating that feature engineering is often the most indispensable part of human intervention in ML pipelines since human intuition and experience are gravely required, thus, it becomes tedious, task-specific and challenging, and hence, time-consuming. 
On the other hand, with the growing need for ML techniques in industrial tasks, it becomes impracticable to manually perform feature engineering in all of these tasks.
This promotes the birth of automatic feature engineering, which is an important topic of automatic machine learning (AutoML) \cite{DBLP:journals/corr/abs-1810-13306,hutter2019automated,DBLP:journals/corr/abs-1904-12054,DBLP:journals/corr/abs-1908-00709}.
The development of automatic feature engineering can not only liberate machine learning engineers from the substantial and tedious process, but also power machine learning techniques to be applied in more and more applications.

For a regular supervised learning task, problem can be formulated as using training examples to find a function $\mathcal{F}: X \rightarrow Y$, which is defined as returning the $y$ value that obtains the highest score: $\mathcal{F}(\bm{x})=\arg\max\limits_{y}\mathcal{S}(\bm{x},y)$, where $X$ is the input space, $Y$ is the output space and $\mathcal{S}: X \times Y \rightarrow R$ is a scoring function.
The goal of automatic feature engineering is to learn a feature representation $\Psi: X \rightarrow Z$, to construct a new feature representation $\bm{z}$ from the original feature $\bm{x}$, with which the performance of subsequent machine learning tools can be further improved as much as possible.

Several studies have been conducted on this topic. 
To name a few, some methods use reinforcement learning based strategy to perform automatic feature engineering~\cite{DBLP:conf/icml/GaudelS10, DBLP:conf/aaai/KhuranaST18, DBLP:conf/atal/ZhangHFW19}. These methods require many rounds of attempts and it is necessary to generate a new feature set and evaluate it in each round, making them infeasible in industrial tasks. 
Transfer learning or meta-learning based strategies are also proposed for automatic feature engineering~\cite{DBLP:conf/icdm/KatzSS16, DBLP:conf/ijcai/NargesianSKKT17}. However, a large number of experiments on various datasets are needed in advance to train these methods, and it is intractable to introduce new operators or increase the number of parent features. 
Some methods follow the generation-selection procedure~\cite{DBLP:conf/dsaa/KanterV15, DBLP:journals/corr/LamTSCMA17, DBLP:conf/icdm/KaulMP17} to do automated feature engineering. However, these methods always perform as generating all legal features in the feature generation stage and then selecting a subset features from them, thus the time and space complexity is extremely high, making it unapplicable for tasks with large data size or high feature dimension.

In industrial tasks, the size of real business data is always very huge, which introduces extremely high requirements for space and time complexity. At the same time, due to the rapid change of business, there are also high requirements for the flexibility and scalability of the algorithms. Besides, there are more requirements that need to be addressed~\cite{DBLP:conf/kdd/LuoWZYTCDY19}, ~\cite{DBLP:journals/tist/ZhangZZFLLLZCLQ19}:
\begin{itemize}
	\item Strong applicability: A tool that is highly adaptable means that it is user-friendly and easy to use. The performance of an automatic feature engineering algorithm should not depend on a large number of hyper-parameters or one of its hyper-parameter configurations can be applied to different data sets.
	\item Distributed computing: the number of samples and features in real-world business tasks are pretty large, which makes distributed computing necessary. Most parts of the automatic feature engineering algorithm should be able to be calculated in parallel.
	\item Real-time inference: real-time inference is involved in many real-world businesses. In such cases, once an instance is inputted, the feature should be produced instantly and the prediction can be performed subsequently.
\end{itemize}

In this paper, we approach the problem from the typical two-stage perspective and propose a method named SAFE (\textbf{S}calable \textbf{A}utomatic \textbf{F}eature \textbf{E}ngineering) to perform efficient automatic feature engineering, which includes feature generation stage and feature selection stage.
We guarantee computational efficiency, scalability and the requirements mentioned above. The major contributions of this paper are summarized as follows:
\begin{itemize}
	\item In the feature generation stage, different from the previous methods which focuses on what operator to use or how to generate all legal features, we focus on mining the original feature pairs that generate more effective new features with higher probability, to improve the efficiency of the process.
	\item In the feature selection stage, we propose a pipeline of feature selection, with the consideration of the power of a single feature, the redundancy of feature pairs, and the feature importance evaluated by the typical tree model. It is suitable for multiple different business data sets and various machine learning algorithms.
	\item We have experimentally proved the advantages of our algorithm on a large set of data sets and multiple classifiers. Compared with the original feature space, the prediction accuracy is improved by $6.50\%$ on average.
\end{itemize}

The rest of this paper is organized as follows: Section~\ref{Related Work} reviews the related work; Section~\ref{Problen Statement} explains the problem setting; Section~\ref{Proposed Method} details the proposed method and provides some analyses; Section~\ref{Experiments} states the detail of the data set, evaluation method and presents the experimental results to validate our method; Section~\ref{Conclusion} concludes the paper.

\section{Related Work}
\label{Related Work}
As a nonnegligible issue for automatic machine learning~\cite{DBLP:journals/corr/abs-1810-13306,hutter2019automated,DBLP:journals/corr/abs-1904-12054,DBLP:journals/corr/abs-1908-00709}, automatic feature engineering has drawn extensive attention in recent years, and many methods have been proposed from different perspectives to solve this task~\cite{DBLP:journals/ml/MarkovitchR02, DBLP:conf/icml/GaudelS10,DBLP:conf/dsaa/KanterV15,DBLP:conf/icdm/KatzSS16,DBLP:journals/eswa/PiramuthuS09,DBLP:conf/sdm/FanZPVZRYY10,DBLP:journals/corr/LamTSCMA17,DBLP:conf/icdm/KaulMP17,DBLP:conf/aaai/KhuranaST18,DBLP:conf/atal/ZhangHFW19,DBLP:conf/ijcai/NargesianSKKT17}. 
In this section, we mainly discuss three typical strategies, which include the generation-selection strategy, reinforcement learning based strategy and transfer learning based strategy.

Given a supervised learning data set, a typical method for automatic feature engineering is to follow the generation-selection procedure. 
The FICUS algorithm~\cite{DBLP:journals/ml/MarkovitchR02} initializes by constructing a set of candidate features, and iterates to improve it until the computation budget is exhausted. During each iteration, it performs beam search to construct new features and selects features typically by using heuristic measures based on information gain in a decision tree.
TFC~\cite{DBLP:journals/eswa/PiramuthuS09} also solves this task by an iterative framework. In each iteration, it generates all legal features based on the current feature pool and all available operators, then selects the best features from all candidate features by using information gain, and keeps them as the new feature pool. With this framework, higher-order feature combinations can be obtained as the iteration progresses. However, the exhaustive search in each iteration leads to a combinatorial explosion of feature space, making this approach non-scalable. 
To avoid exhaustive search, learning based methods, such as the FCTree algorithm \cite{DBLP:conf/sdm/FanZPVZRYY10}, have been proposed. FCTree trains a decision tree and performs feature generation by applying several sequential transformations to the original feature, and select features according to information gain on each node of the decision tree. Once a tree is built, features chosen at internal decision nodes are used to obtain the constructed features.
\cite{DBLP:conf/icdm/KaulMP17} is a regression-based algorithm, which learns the representation by mining pairwise feature associations, identifying the linear or non-linear relationship between each pair, applying regression and selecting those relationships that are stable and improve the prediction performance. 
These algorithms always encounter performance and scalability bottlenecks since the cost of time and resource in the feature generation and selection procedure may be extremely unsatisfactory, if without ingenious design.

Reinforcement learning based strategies are also explored. \cite{DBLP:conf/icml/GaudelS10} formalizes feature selection as a reinforcement learning problem and introduces an adaptation of the Monte-Carlo tree search. Here, the problem of choosing a subset of the available features is cast as a single-player game whose states are all possible subsets of features and the actions consist of choosing a feature and adding it to the subset.
\cite{DBLP:conf/aaai/KhuranaST18} handles this problem by exploring on a directed acyclic graph which represents the relationship between different transformed versions of the data, and learns an effective strategy to explore available feature engineering choices under a given budget through Q-learning.
\cite{DBLP:conf/atal/ZhangHFW19} formalizes this task as an optimization problem over a Heterogeneous Transformation Graph (HTG). It proposes a deep Q-learning on HTG to support efficient learning of fine-grained and generalized FE policies that can transfer knowledge of engineering ``good" features from a collection of data sets to other unseen data sets.

Transfer learning or meta-learning based strategies are also proposed for automatic feature engineering. \cite{DBLP:conf/icdm/KatzSS16} employs learning to rank techniques to evaluate the newly constructed features and select the most promising ones. It is extremely time-consuming. For instance, their reported results were obtained after running for several days on moderately sized data sets. 
In contrast, \cite{DBLP:conf/ijcai/NargesianSKKT17} can generate effective features within seconds on average. It is based on learning the effectiveness of applying a transformation (e.g., arithmetic or aggregate operators) on numerical features, from past feature engineering experiences. However, since the meta-features do not take the relationship between features into account, it works better only when using unary transformations.

Beyond that, there are also other methods that direct at different settings. For example, \cite{DBLP:conf/dsaa/KanterV15} automatically constructs features from relational databases via deep feature synthesis. It focuses on the relationships between the various tables in the database to generate new features. A similar approach is adopted by \cite{DBLP:journals/corr/LamTSCMA17}. 
What's more, many studies try to perform feature engineering simultaneously while training the model, by introducing operations such as feature cross~\cite{DBLP:conf/kdd/WangFFW17} or using techniques like self-attentive neural networks~\cite{DBLP:conf/cikm/SongS0DX0T19}. 
We need to address that, different from the methods that learn feature representations simultaneously with model training, we are aiming at learning a new representation for each sample based on the original features, which can be used to perform the subsequent machine learning models, and we have no constraint on what model to be used afterward. 
At the same time, for industrial tasks, interpretability is always required~\cite{DBLP:conf/cikm/ZhangL19}. The generated features in our framework can be easily explained, to satisfy the interpretability requirement in industrial tasks.

To apply automatic feature engineering techniques in real-world applications, especially for industrial tasks, the efficiency and scalability of the aforementioned methods are still far from satisfactory. Methods with excellent efficiency and scalability, along with requisite interpretability and promising performance are in high demand.

\section{Problem Statement}
\label{Problen Statement}

Consider a predictive modeling task, which consists of:

\begin{itemize}
	\item A dataset of input-output pairs. Let $\bm{x} \in X$ be a record of the input space with $M$ original features $\left\{\bm{x}^{(1)},\cdots,\bm{x}^{(M)}\right\}$. Let $y \in Y$ be the corresponding output label. Training data with $N$ records can be denoted as $\mathcal{D}_{train}=\left\{X_{train}, Y_{train}\right\}=\left\{(\bm{x}_1,y_1),\cdots,(\bm{x}_N,y_N)\right\}$. Similarly, validation data and test data can be denoted as $\mathcal{D}_{valid}$ and $\mathcal{D}_{test}$.
	\item A machine learning algorithm $\mathcal{A}$ that accepts a training set and a validation set as input and produces a function $\mathcal{F}: X \rightarrow Y$, which return the predicted label $y$ given the input $x$.
	\item A loss function $\mathcal{L}$ which computes the loss of a learned function $\mathcal{F}$, according to the ground-truth label $y$.
\end{itemize}

The goal of automatic feature engineering is to learn a feature generation function $\Psi: X \rightarrow Z$ to generate new feature representation $\bm{z} \in Z $ based on the original features $\bm{x} \in X$, by using the set of $k$ operations $\mathcal{O}=\left\{o_1,\cdots,o_k\right\}$, so that the learning algorithm $\mathcal{A}$ can find a function $\mathcal{F}$ that minimizes the loss function $\mathcal{L}$.
i.e., to approximate the true underlying input-output function as much as possible. More formally, we want to obtain the feature generation function, with which the loss $\mathcal{L}$ of the learned predictive function $\mathcal{F}$ can be minimized:
\begin{equation}
\Psi^* = \arg\min\limits_{\Psi}\mathcal{L}(\mathcal{F}( \Psi(X_{test})), Y_{test})
\end{equation}
in which the predictive function can be obtained by $\mathcal{F}=\mathcal{A}(\mathcal{D}_{train\_new}, \mathcal{D}_{valid\_new})$, $\mathcal{D}_{train\_new}$ and $\mathcal{D}_{valid\_new}$ are the generated training and validation dataset, i.e., $\mathcal{D}_{train\_new}=\left\{ \Psi(X_{train}), Y_{train}\right\}$, and $\mathcal{D}_{valid\_new}=\left\{ \Psi(X_{valid}), Y_{valid}\right\}$.

The operators $\mathcal{O}$, also known as $n$-ary operators, acts on $n$ original features for feature generation, and it can be divided into unary operators $\mathcal{O}_1$, binary operators $\mathcal{O}_2$, ternary operators $\mathcal{O}_3$, etc. It should be noted that operators which do not satisfy the commutative property will be treated as multiple different operators in our subsequent descriptions and experiments, such as ``$\div$''.

Unary operators are used for discretizing, normalizing, or mathematically transforming unit features:
\begin{itemize}
	\item Discretization is the process of transferring continuous features into discrete features. It plays an important role in feature processing. It is robust to anomalous data and can make the trained model more stable. Typical feature discretization methods include ChiMerge, equidistant binning, equal-frequency binning, clustering binning, etc.
	\item Normalization refers to a process that makes features more normal or regular. Typical feature normalization methods include Min-Max normalization, Z-score, standardization of dispersion, etc.
	\item Mathematical transformations acting on unit features include log, sigmoid, square, square root, tanh, round, etc.
\end{itemize}

Binary operators combine two original features to generate a new feature:
\begin{itemize}
	\item Four basic arithmetic operations: $+$, $-$, $\times$, $\div$.
	\item Logical operators act on two boolean features, such conjunction ($\wedge$), disjunction ($\vee$), alternative denial ($\uparrow$), joint denial ($\downarrow$), material conditional ($\rightarrow$), converse implication ($\leftarrow$), biconditional ($\leftrightarrow$), exclusive or ($\nleftrightarrow$), etc. 
	\item \textit{GroupByThenMax}, \textit{GroupByThenMin}, \textit{GroupByThenAvg}, \textit{GroupByThenStdev} and \textit{GroupByThenCount}. These operators implement the SQL-based operations with the same name.
	\item Ridge regression and kernel ridge regression in \cite{DBLP:conf/icdm/KaulMP17} can also be considered as binary operators.
\end{itemize}

Ternary operators combine three features to generate a new feature. A common ternary operator is a conditional operator, which is a basic conditional statement in many programming languages. For the conditional expression $a?b:c$, if the value of $a$ is true, the value of $b$ is obtained; otherwise, the value of $c$ is obtained.

There are also many operators that can accept multiple original features as input, such as MAX, MIN, MEAN, etc. We divide them into different categories when they accept a different number of inputs.

It should be pointed out that there are still many operators that apply in specific fields, we call them domain-specific operators, such as lag operators in time series analysis, genetic operators in biology, etc.

Because of the existence of various operators, an applicable automatic feature engineering algorithm framework should not limit operators and new operators should be easily added.


What's more, to ensure the method to be feasible for large scale industrial tasks, the whole automatic feature engineering framework should be time and space-friendly, and with requisite interpretability and promising performance.

\section{Proposed Method}
\label{Proposed Method}

\subsection{Overview}

\begin{figure}[htbp]
	\centerline{\includegraphics[width=3.2in]{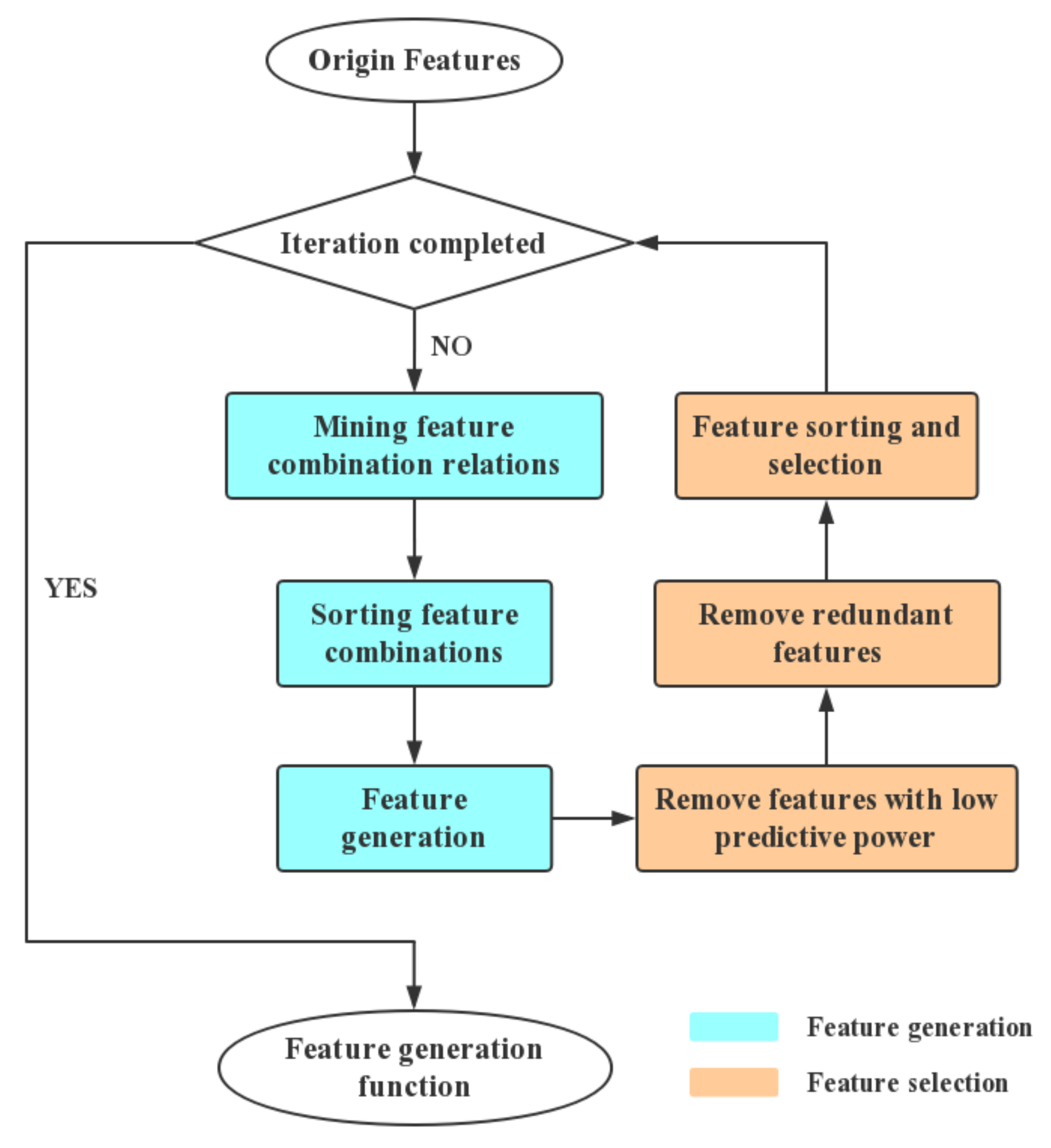}}
	\caption{Flow chart of SAFE}
	\label{SAFE}
\end{figure}

As shown in Fig. \ref{SAFE} and Algorithm \ref{Pseudo-code of SAFE}, our automatic feature engineering algorithm is an iterative process where each iteration comprises of two phases: feature generation and feature selection. The number of iterations is limited by the calculation time or computation space.

\begin{algorithm*}[htbp]
	\caption{Pseudo-code of proposed method SAFE}
	\label{Pseudo-code of SAFE}
	\begin{algorithmic}[1]
		\Require
		The raw data set $D_{train}$ and $D_{valid}$;
		The operators $\mathcal{O}$;
		Number of iterations $nIter$ or iteration time $tIter$.
		\Ensure
		Feature generation function $\Psi$.
		\State $i \leftarrow 0, t_{start} \leftarrow now()$
		\While{$i < nIter$ and $now()-t_{start} < tIter$}
		\State Train XGBoost model on $D_{train}$ and $D_{valid}$.
		\State Constitute feature combinations from the same path of the model.
		\State Sort and filter feature combinations to get $\widetilde{P}$ by information gain ratio. See Algorithm \ref{Pseudo-code of sorting feature combinations}.
		\State Apply the operations on the filtered feature combinations $\widetilde{P}$ to get the generated features $\widetilde{X}$.
		\State Construct candidate features $\hat{X}^i$ by combining the generated features $\widetilde{X}$ and the base features $X$ in this iteration.
		\State Remove features with low predictive power from $\hat{X}$, and get $\hat{X}^A$. See Algorithm \ref{Pseudo-code of removing features with low predictive power}.
		\State Remove redundant features $\hat{X}^A$, and get $\hat{X}^B$. See Algorithm \ref{Pseudo-code of removing redundant features}.
		\State Sort the remaining candidate features $\hat{X}^B$ and select the features with high importance to form $\hat{X}^C$
		\State $D_{train} \leftarrow \left\{\hat{X}_{train}^C, Y_{train}\right\}$, $D_{valid} \leftarrow \left\{\hat{X}_{valid}^C, Y_{valid}\right\}$
		\State $i \leftarrow i+1$
		\EndWhile
		\State The feature generation function $\Psi$ is obtained from the selected features in the last iteration.\\
		\Return $\Psi$.
	\end{algorithmic}
\end{algorithm*}

As discussed above, exhaustive searching is ineffective due to the infinite feature space. Even if the numbers of operators and iterations are limited, exhaustive searching can also result in a combinatorial explosion. 
To avoid this problem, we use a tree based method, i.e., XGBoost \cite{DBLP:conf/kdd/ChenG16}, to mine the relationships between the current set of base features $X^i$ to narrow down the search space for feature combinations and then sort and filter the feature combinations by information gain ratio. 
We then apply the predefined operators on the filtered feature combinations with a high information gain ratio and obtain the new feature set $\widetilde{X}^i$. By combining the base features $X^i$ and the generated features $\widetilde{X}^i$, the candidate feature set can be obtained, which is denoted as $\hat{X}^i = X^i \cup \widetilde{X}^i$.

As the number of the current set of features $\hat{X}^i$ is still very large, we use efficient and effective feature ranking and selection methods after that. 
The basic idea is to find the informative features, remove the redundant ones, and then each feature with be attached a score so that the filter process can be performed if necessary. 
Concretely, we first use the information value to pick out the features that have a high impact on the label, which are regarded as more informative features. 
Then, we use the pearson correlation coefficient to remove the redundant features. 
Finally, we use XGBoost to score the remaining features by the average gain across all splits in which the feature is used. 
We only choose the features with the highest scores as $X^{i+1}$ for the next iteration, with the consideration of scalability and efficiency.

In the next two subsections, we will explain the feature generation and feature selection process in detail.

\subsection{Feature Generation}

The goal of this phase is to ingeniously generate the set of the new feature set $\widetilde{X}^i$ using the current feature set $X^i$. 
Moreover, we want to reduce the number of newly generated features, while the effective ones should not be omitted.
The training set, validation set, and test set at this point are represented as $D^i_{train}$, $D^i_{valid}$ and $D^i_{test}$.

The search space of original feature generation is:
\begin{equation}
\small
\mathcal{S}=\bigcup\limits_{i=1}^{M}\left\{\left\{\bigcup\limits_{1 \leq s_1 \leq \cdots \leq s_i \leq M}\left\{x^{s_1}, \cdots, x^{s_i}\right\}\right\} \times \mathcal{O}_i\right\}
\end{equation}
where $M$ is the number of original features and $\mathcal{O}_i$ represents the set of $i$-ary operators.

The number of elements in the search space is:
\begin{equation}
\mathcal{T}=\sum\limits_{i=1}^{M}(\mathcal{A}_{M}^i\times|\mathcal{O}_i|)
\end{equation}
where $\mathcal{A}^k_n$ represents the number of ways of obtaining an ordered subset of $k$ elements from a set of $n$ elements.

\begin{figure}[htbp]
	\centerline{\includegraphics[width=3.2in]{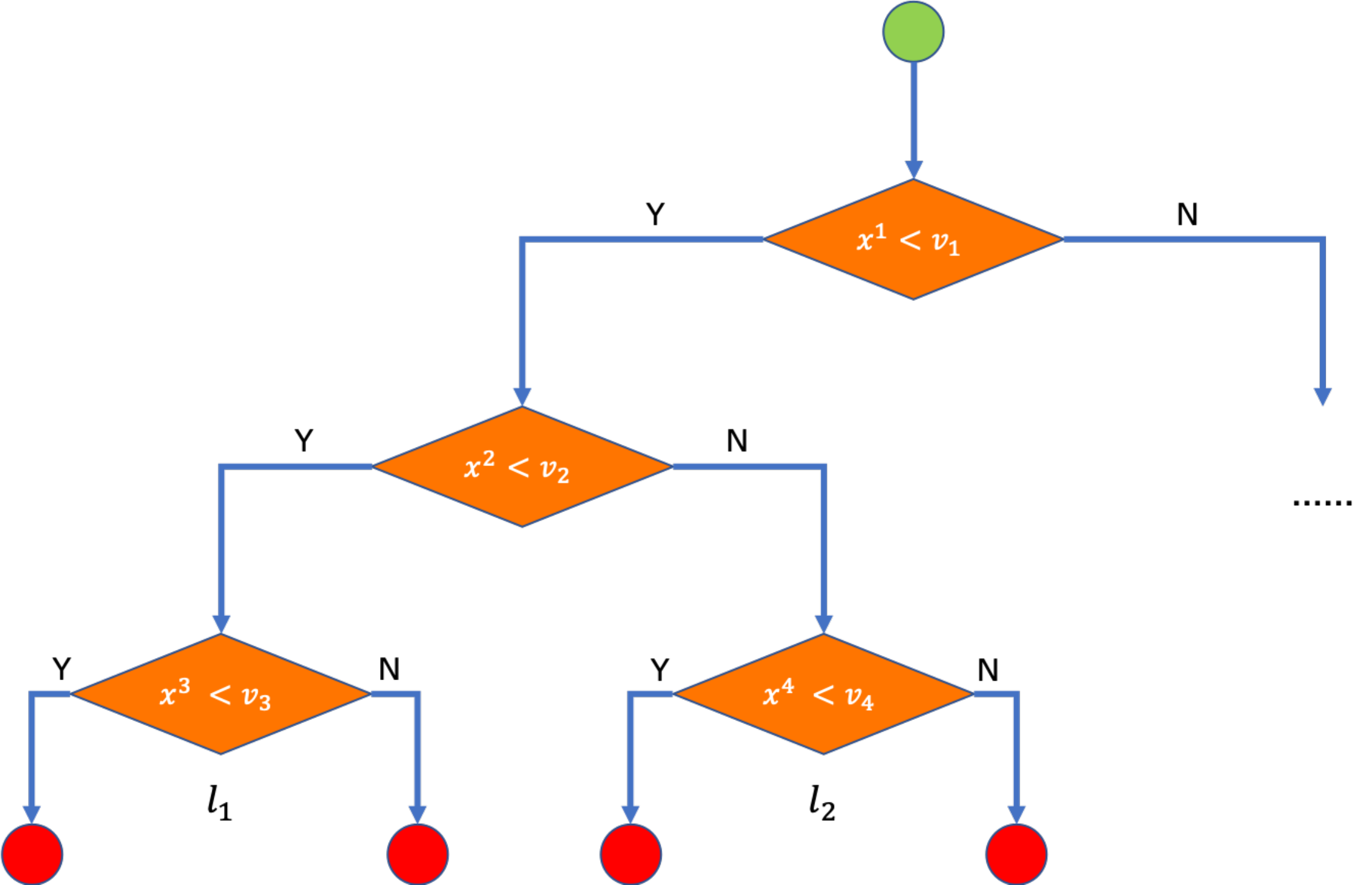}}
	\caption{Example of a regression tree in the XGBoost model}
	\label{regression_tree}
\end{figure}

\subsubsection{Mine feature combination relations}
As mentioned earlier, this search space is so large that we have to narrow it down, and the informative feature combinations should not be ignored. 
First we train a tree model, i.e., XGBoost, on $D^i_{train}$ and $D^i_{valid}$. Consider a regression tree in the XGBoost model, as shown in Fig.~\ref{regression_tree}. We call $\left\{\bm{x}^{(i)}\right\}$ as the split features and $\left\{v_i\right\}$ as the corresponding split values, and the features which do not act as a split feature are called non-split features. The parent node of the leaf node is represented as $l_j$ and the different split features on a path of the tree from the root node to $l_j$ can be represented as $p_j$. For example, $p_1=\left\{\bm{x}^{(1)}, \bm{x}^{(2)}, \bm{x}^{(3)}\right\}$ and $p_2=\left\{\bm{x}^{(1)}, \bm{x}^{(2)}, \bm{x}^{(4)}\right\}$ in Fig.~\ref{regression_tree}. All $k$ paths of all trees in the XGBoost model can be represented as $P=\left\{p_1, \cdots, p_k\right\}$. Our automatic feature engineering algorithm SAFE is based on two basic assumptions:

\begin{itemize}
	\item For unary operators, features that generated based on split features are more efficient than that generated based on non-split features.
	\item For other operators, new features that generated based on split features which from the same path are more efficient than the new features that generated based on split features which from different path, while the latter is still more efficient than features that generated based on non-split features.
\end{itemize}

We empirically verify the rationality of these two assumptions in section \ref{Experiments}. Based on these two basic assumptions, we can find features or feature combinations on these paths for feature generation, which not only greatly reduces the search space, guarantees efficiency, but also ensures the validity of feature generation. The search space of feature generation by this way is:
\begin{equation}
\small
\mathcal{S}^*=\bigcup\limits_{i=1}^{k}\bigcup\limits_{j=1}^{|p_i|}\left\{\left\{\bigcup\limits_{1 \leq s_1 \leq \cdots \leq s_j \leq |p_i|}\left\{p^{s_1}_i, \cdots, p^{s_j}_i\right\}\right\} \times \mathcal{O}_j\right\}
\end{equation}
where $k$ is the number of paths and $\mathcal{O}_i$ represents the set of $i$-ary operators.

The maximum number of elements in the search space is:
\begin{equation}
\mathcal{T}^*=\sum\limits_{i=1}^{k}\sum\limits_{j=1}^{|p_i|}(\mathcal{A}_{|p_i|}^j\times|\mathcal{O}_j|)
\end{equation}
where $p_i^j$ means the $j$-th feature on the path. It should be noted that some combinations of features on different paths may be the same, so the actual number will be much smaller than this value. It can be found through formulas and experiments that: $\mathcal{T}^*<<\mathcal{T}$.

\subsubsection{Sort feature combinations}
To further narrow down the search space of feature generation, we use the information gain ratio to sort the features and feature combinations in the search space. Take a feature combination with $q$ elements $\left\{\bm{x}^{(1)},\cdots,\bm{x}^{(q)}\right\}$ as an example, we already know their split values $V_1,\cdots,V_q$. Here, $V_i$ is a collection because a split feature may appear multiple times in a path. These split features and split values can divide all records into $\prod\limits_{i=1}^q(|V_i|+1)$ parts. The information gain ratio can be calculated by subtracting the original information entropy from the information entropy on these parts. The pseudo-code of the algorithm at this phase is shown in Algorithm \ref{Pseudo-code of sorting feature combinations}.

\begin{algorithm}[htbp]
	\caption{Sorting feature combinations}
	\label{Pseudo-code of sorting feature combinations}
	\begin{algorithmic}[1]
		\Require
		All paths in the XGBoost model $P$;
		Number of output features or feature combinations $\gamma$.
		\Ensure
		$\gamma$ features or feature combinations $\widetilde{P}$.
		\For{each path $p^i$ in $P$}
		\For{each combination $c_j=\left\{x^{(s_1)},\cdots,x^{(s_q)}\right\}$ in $p^i$}
		\State Divide all records into $\prod\limits_{i=1}^q(|V_{s_i}|+1)$ parts according to $\left\{x^{(s_1)},\cdots,x^{(s_q)}\right\}$ and $\left\{V_{s_1},\cdots,V_{s_q}\right\}$.
		\State Calculate the information gain raito of $c_j$.
		\EndFor
		\EndFor
		\State Select the $\gamma$ features or feature combinations with the highest information gain ratio and mark them as $\widetilde{P}$.\\
		\Return $\widetilde{P}$.
	\end{algorithmic}
\end{algorithm}

\subsubsection{Generate features}
\label{Feature generation}

$\gamma$ features or feature combinations with the highest information gain ratio will be used to generate $\sum\limits_{i=1}^M(\gamma_i \times |\mathcal{O}_i|)$ new features $\widetilde{X}^i$, where $\gamma_i$ denotes the number of feature combinations with $i$ features.

Since the number of features that searched and generated is much less than exhaustive searching, this allows us to employ iterative feature generation strategies on large data sets.

\subsection{Feature Selection}
\label{Feature Selection and Ranking}

Candidate features $\hat{X}^i$ that composed of the base features $X^i$ in this iteration and generated features $\widetilde{X}^i$ might not be of equal importance. To computationally-efficiently select the more informative features, we use a three-step feature selection process: Firstly, according to the information value, the features with low predictive power are removed. Then the redundant features are removed according to the pearson correlation coefficient. Finally, the remaining features are sorted by using a tree based method, i.e., XGBoost \cite{DBLP:conf/kdd/ChenG16}.

\subsubsection{Remove uninformative features}

Since some of the candidate features will inevitably have little or no impact on the target, we first remove the features with low predictive power. The pseudo-code of the algorithm at this phase is shown in Algorithm \ref{Pseudo-code of removing features with low predictive power}.

\begin{algorithm}[htbp]
	\caption{Remove features with low predictive power}
	\label{Pseudo-code of removing features with low predictive power}
	\begin{algorithmic}[1]
		\Require
		The data set $D_{train}^*=\left\{\hat{X}_{train},Y_{train}\right\}$;
		Threshold of information value $\alpha$;
		Number of bins $\beta$.
		\Ensure
		Selected candidate feature set $\hat{X}_{train}^A$.
		\State $\hat{X}_{train}^A \leftarrow \varnothing$
		\For{each candidate feature $x^{(i)}$ in $\hat{X}_{train}$}
		\State Pack $D_{train}^*$ into $\beta$ bins at the same frequency.
		\State Calculate $IV_i$, the IV of feature $x^{(i)}$ by Eq.~(\ref{Information value}).
		\If{$IV_i > \alpha$}
		\State $\hat{X}_{train}^A = \hat{X}_{train}^A \cup \left\{x^i\right\}$.
		\EndIf
		\EndFor\\
		\Return $\hat{X}_{train}^A$.
	\end{algorithmic}
\end{algorithm}

Information value (IV) is a very useful concept for feature selection during model building, and it is widely used in the industrial tasks. IV measures the degree to which a feature affects the target. The formula for information value is shown below:
\begin{equation}
\label{Information value}
IV=\sum\limits_{i}(\frac{n_p^i}{n_p}-\frac{n_n^i}{n_n})\times\frac{n_p^i/n_p}{n_n^i/n_n}
\end{equation}
where $n_p$ and $n_n$ represent the number of all positive records and negative records. $n_p^i$ and $n_n^i$ represent the number of positive records and negative records in the $i$-th bin.

\begin{table}[htbp]
	\caption{Information Value}
	\begin{center}
		\setlength{\tabcolsep}{6mm}{
			\begin{tabular}{c|c}
				\hline
				Information Value & Predictive Power\\
				\hline
				$0$ to $0.02$ & Useless for prediction\\
				$0.02$ to $0.1$ & Weak predictor\\
				$0.1$ to $0.3$ & Medium predictor\\
				$0.3$ to $0.5$ & Strong predictor\\
				$>0.5$ & Extremely strong predictor\\
				\hline
			\end{tabular}
		}
		\label{Information Value}
	\end{center}
\end{table}

As shown in Table \ref{Information Value}, the rules of thumb guide us on how to remove features with low predictive power. Typically, variables with medium and strong predictive powers are selected for model development. Therefore, we take the threshold of feature selection as $\alpha = 0.1$.

\subsubsection{Remove redundant features}

The candidate features at this time are with certain predictive power, but some of them are redundant. For example, ``speed'' and ``one-hour travel'' are highly relevant, we only need to keep one. The pseudo-code of the algorithm at this phase is shown in Algorithm \ref{Pseudo-code of removing redundant features}.

\begin{algorithm}[htbp]
	\caption{Remove redundant features}
	\label{Pseudo-code of removing redundant features}
	\begin{algorithmic}[1]
		\Require
		The data set $D_{train}^*=\left\{\hat{X}_{train}^A,Y_{train}\right\}$;
		Threshold of pearson correlation $\theta$.
		\Ensure
		Selected candidate feature set $\hat{X}_{train}^B$.
		\State $\hat{X}_{train}^B \leftarrow \varnothing$
		\For{each candidate feature $x^{(i)}$ in $\hat{X}_{train}^A$}
		\For{each candidate feature $x^{(j)}$ in $\hat{X}_{train}^A$}
		\If{$i<j$}
		\State Calculate $Pearson(i,j)$ by Eq.~(\ref{Pearson correlation}).
		\If{$|Pearson(i,j)|> \theta$}
		\If{$IV_i>IV_j$}
		\State $\hat{X}_{train}^B=\hat{X}_{train}^B \cup \left\{x^{(i)}\right\}$
		\Else
		\State $\hat{X}_{train}^B=\hat{X}_{train}^B \cup \left\{x^{(j)}\right\}$
		\EndIf
		\EndIf
		\EndIf
		\EndFor
		\EndFor\\
		\Return $\hat{X}_{train}^B$.
	\end{algorithmic}
\end{algorithm}

A pearson correlation is a number between $-1$ and $1$ that indicates the extent to which two features are linearly related. Its absolute value of 1 means that the two features are completely linearly related, and its absolute value of 0 means there is no linear relationship between the two features. $Pearson(i,j)$, a pearson correlation between features $\bm{x^{(i)}}$ and $\bm{x^{(j)}}$ is calculated by:
\begin{equation}
\small
\label{Pearson correlation}
Pearson(i,j)=\frac{\sum\limits_{k=1}^N(\bm{x^{(i)}_k}-\overline{\bm{x^{(i)}}})(\bm{x^{(j)}_k}-\overline{\bm{x^{(j)}}})}{\sqrt{\sum\limits_{k=1}^N(\bm{x^{(i)}_k}-\overline{\bm{x^{(i)}}})^2}\sqrt{\sum\limits_{k=1}^N(\bm{x^{(j)}_k}-\overline{\bm{x^{(j)}}})^2}}
\end{equation}
where $\overline{\bm{x^{(i)}}}$ and $\overline{\bm{x^{(j)}}}$ means the average of all elements of feature $i$ and feature $j$.

The larger the absolute value of the correlation coefficient, the stronger the correlation is. Usually, the relative strength of the variable is judged by the range of values in Table \ref{Pearson Correlation Coefficient}. Therefore, we set the threshold $\theta$ of pearson correlation to $0.8$. If the pearson correlation coefficient of the two features is greater than $0.8$, the feature with the smaller IV of them will be removed.

\begin{table}[htbp]
	\caption{Pearson Correlation}
	\begin{center}
		\setlength{\tabcolsep}{4mm}{
			\begin{tabular}{c|c}
				\hline
				Pearson Correlation Coefficient & Correlation\\
				\hline
				$0$ to $0.2$ & Very weak or no correlation\\
				$0.2$ to $0.4$ & Weak correlation\\
				$0.4$ to $0.6$ & Moderate correlation\\
				$0.6$ to $0.8$ & Strong correlation\\
				$0.8$ to $1$ & Extremely strong correlation\\
				\hline
			\end{tabular}
		}
		\label{Pearson Correlation Coefficient}
	\end{center}
\end{table}

\subsubsection{Rank feature importance}

At this stage, we use a lightweight tree-based method, i.e., XGBoost, to sort the remaining candidate features by the average gain across all splits, and further filter can be performed if a maximum value of the number of final selected features is required to make the later process more efficient.

\subsection{Time complexity Analysis}
\label{Time complexity}

In this section, we analyze the time complexity of the algorithm. We first analyze the time complexity of some existing algorithms. 
Reinforcement learning based strategies are beyond our consideration since the executing time of them is too long. 
We mainly analyze generation-selection based strategies (TFC, FCTree, AutoLearn) and transfer learning or meta-learning based strategies (ExploreKit, LFE). 
Then we analyze the time complexity of SAFE. 
It should be noted that for the sake of simplicity and generality, we only consider the first iteration of all iterative algorithms (TFC, ExploreKit, SAFE), and we only consider binary operators. Recall that we denote the number of records and features as $N$ and $M$, respectively, and $\mathcal{A}^k_n$ represents the number of ways to obtain an ordered subset of $k$ elements from a set of $n$ elements.

\subsubsection{TFC}

TFC \cite{DBLP:journals/eswa/PiramuthuS09} generates all legal features and then selects the best ones using information gain. The time complexity of feature generation is $O(N\mathcal{A}_M^2)=O(NM^2)$, the time complexity of feature selection is $O(NM^2)$, the time complexity of feature ranking is $O(M^2\log M^2)$. For real business data with a large amount of data, $\log M$ is always much smaller than $N$, so its time complexity is:
\begin{equation}
O_{TFC}=O(M^2(N+\log M^2))=O(NM^2)
\end{equation}

\subsubsection{FCTree}

The time complexity of decision tree algorithm is $O(NMD) \leq O(NM\log N)$, in which $D$ is the depth of the tree. FCTree \cite{DBLP:conf/sdm/FanZPVZRYY10} algorithm adds $n_e$ features at each level of decision tree, so its time complexity is $O(NM\log N+\frac{1}{2}N(n_e(\log N-1))*\log N)=O((M+n_e\log N)N\log N)$. For real business data with large amount of data, $M$ is always much smaller than $\log N$, so its time complexity is:
\begin{equation}
O_{FCTree}=O(n_eN(\log N)^2)
\end{equation}

\subsubsection{AutoLearn}

AutoLearn \cite{DBLP:conf/icdm/KaulMP17} identifies the linear or non-linear relationship between each pair and uses randomized lasso and mutual information for feature selection. The time complexity of feature generation is $O(M^2) \times (O_{Ridge}+O_{KernelRidge})=O(NM^2)$, the time complexity of feature selection is $O_{Lasso}+O_{MI}=O_{Lasso}+O(NM^2)$, the time complexity of feature ranking is $O(M^2\log M^2)$. So the time complexity is:
\begin{equation}
O_{AutoLearn}=O(M^2(N+\log M^2))+O_{Lasso}
\end{equation}

\subsubsection{ExploreKit}

ExploreKit \cite{DBLP:conf/icdm/KatzSS16} is a meta-learning based strategies. In the feature generation phase, it needs an exhaustive combination of features, so its time complexity of feature generation is $O(NM^2)$. In the feature ranking phase, it calculates the meta-features associated with the original data set and candidate features, such as entropy-based measures and statistical tests, to score each candidate feature, so its time complexity of feature generation is $O(NM^2)+M^2 \times O_{score}$. In addition, it needs $O_{Meta}$ to train a meta-learning model in advance. So the all-time complexity is:
\begin{equation}
O_{ExploreKit}=O_{Meta} + O(NM^2)+M^2 \times O_{score}
\end{equation}

\subsubsection{LFE}

LFE \cite{DBLP:conf/ijcai/NargesianSKKT17} is also a meta-learning based strategy. The difference is that it does not require exhaustive feature generation. At the feature selection stage, meta-features are only related to the original features. So the whole complexity can be calculated as:
\begin{equation}
O_{LFE}=O_{Meta} + O(NM)+M^2 \times O_{score}
\end{equation}

\subsubsection{SAFE}

The most important calculation at the phases of mining feature combination relations and feature importance ranking is to train an XGBoost. Their time complexity is $O(NMK_1D_1+NM\log R)$ and $O(N\hat{M}^BK_2D_2+N\hat{M}^B\log R)$, respectively. Where $K_1$ and $K_2$ mean the total number of trees, $D_1$ and $D_2$ mean the maximum depth of the tree and $R$ is the maximum number of rows in each block \cite{DBLP:conf/kdd/ChenG16} and $\hat{M}^B$ is the number of features after removing redundant features. The number of features after feature generation and removing uninformative features can be denoted as $\hat{M}$ and $\hat{M}^A$, respectively. Next, we analyze the time complexity of the other four phases:

\begin{itemize}
	\item Sorting feature combinations: As shown in Algorithm \ref{Pseudo-code of sorting feature combinations}, there are $2^{D_1}K_1\mathcal{A}_{D_1}^2$ binary feature combinations and the time complexity of this phase is $O(2^{D_1}K_1N{D_1}^2)$.
	\item Feature generation: As shown in section \ref{Feature generation}, $\gamma_2 \times |\mathcal{O}_2|$ new features will be generated, so the time complexity of this phase is $O(\gamma_2N|\mathcal{O}_2|)=O(\gamma_2N)$.
	\item Remove uninformative features: As the Algorithm \ref{Pseudo-code of removing features with low predictive power} shows, the time complexity of the third and fourth steps is $O(N)$ and $O(N)$, respectively. So the overall time complexity of this phase is $O(\hat{M}N)$.
	\item Remove redundant features: As the Algorithm \ref{Pseudo-code of removing redundant features} shows, pearson correlation is calculated once for each feature pair. Because the time complexity of Pearson correlation calculation is $O(N)$, the overall time complexity of this phase is $O(N(\hat{M}^A)^2)$.
\end{itemize}

The trees in XGBoost are usually not deep, so we can treat $D$ as a constant and ignore it. For real business data with large amount of data, $\log R<\log N<M<\hat{M}_C<\hat{M}^B<\hat{M}^A<\hat{M}<<N$, $\hat{M}=\gamma_2|\mathcal{O}_2|$ and $\log R<K_1,K_2$. So the all time complexity is:
\begin{equation}
\label{Time complexity of SAFE}
\begin{aligned}
O_{SAFE}&=O(N(\hat{M}^A)^2)+O(N\hat{M}^BK_2)\\
&\leq O(N\hat{M}(\hat{M}+K_2))=O(N\gamma_2(\gamma_2+K_2))\\
&\leq O(N2^{D_1}K_1\mathcal{A}_{D_1}^2(2^{D_1}K_1\mathcal{A}_{D_1}^2+K_2))\\
&=O(NK_1(K_1+K_2))\\
\end{aligned}
\end{equation}

As shown in Eq.~(\ref{Time complexity of SAFE}), we can easily control the number of features generated and the time complexity of the algorithm by controlling the total number of trees of XGBoost.

\begin{table*}[htbp]
	\caption{Classification performance on benchmark data sets}
	\label{Classification performance on benchmark data sets}
	\begin{center}
		\scalebox{0.98}{
			\begin{tabular}{cccccccc|cccccccc}
				\hline
				Dataset & CLF & ORIG & FCT & TFC & RAND & IMP & SAFE & Dataset & CLF & ORIG & FCT & TFC & RAND & IMP & SAFE\\
				\hline
				& AB & 52.10 & 78.27 & 87.33 & 88.20 & 88.30 & \textbf{88.92} & & AB & 99.55 & 99.38 & 96.59 & \textbf{99.62} & 99.58 & 99.53\\
				& DT & 54.38 & 70.78 & 77.37 & 77.37 & 77.43 & \textbf{78.32} & & DT & 98.20 & 98.90 & 96.38 & 98.83 & 98.78 & \textbf{98.99}\\
				& ET & 55.54 & 76.64 & \textbf{88.44} & 85.77 & 86.03 & 88.33 & & ET & 99.83 & 99.76 & 96.95 & 99.88 & 99.88 & \textbf{99.93}\\
				& \textit{k}NN & 51.84 & 78.69 & \textbf{94.81} & 93.01 & 92.69 & 93.66 & & \textit{k}NN & 99.86 & 99.75 & 97.45 & 99.88 & 99.90 & \textbf{99.96}\\
				valley & LR & 58.54 & 80.69 & 92.53 & 93.31 & 93.39 & \textbf{93.80} & banknote & LR & 94.80 & 96.47 & 88.14 & 97.24 & 97.12 & \textbf{97.79}\\
				& MLP & 59.63 & 81.27 & 92.71 & 93.93 & 93.97 & \textbf{94.08} & & MLP & 98.77 & 98.51 & 88.07 & 99.25 & 99.21 & \textbf{99.36}\\
				& RF & 54.54 & 76.31 & 87.06 & 84.34 & 84.34 & \textbf{87.11} & & RF & 99.01 & 99.46 & 96.98 & 99.51 & 99.48 & \textbf{99.57}\\
				& SVM & 65.62 & 81.53 & 93.39 & 94.22 & 94.38 & \textbf{94.94} & & SVM & 98.29 & 98.03 & 88.22 & 98.56 & 98.53 & \textbf{98.64}\\
				& XGB & 54.75 & 86.92 & 94.75 & 95.14 & 95.23 & \textbf{95.68} & & XGB & 99.83 & 99.88 & 99.25 & 99.91 & 99.89 & \textbf{99.91}\\
				\hline
				& AB & 85.15 & 85.24 & 86.22 & 85.60 & 87.54 & \textbf{90.19} & & AB & 93.61 & 93.49 & 92.06 & 93.61 & 93.59 & \textbf{93.74}\\
				& DT & 85.50 & 85.75 & 84.01 & 85.69 & 86.11 & \textbf{87.53} & & DT & 91.28 & 91.06 & 90.72 & 91.51 & 91.65 & \textbf{92.18}\\
				& ET & 91.65 & 91.31 & 88.44 & 91.59 & 92.27 & \textbf{92.79} & & ET & 94.22 & 93.09 & 92.43 & 94.34 & \textbf{94.48} & 94.45\\
				& \textit{k}NN & 83.95 & \textbf{91.98} & 88.46 & 85.03 & 88.03 & 89.08 & & \textit{k}NN & 89.10 & 89.30 & \textbf{91.21} & 90.38 & 90.44 & 91.09\\
				gina & LR & 85.12 & 86.23 & 86.38 & 85.61 & 87.34 & \textbf{90.35} & spambase & LR & 87.71 & 87.86 & 83.68 & 89.48 & 89.53 & \textbf{90.15}\\
				& MLP & 93.09 & 93.81 & 89.09 & 93.21 & 93.75 & \textbf{93.83} & & MLP & 93.81 & 93.43 & 91.04 & 93.84 & 93.81 & \textbf{93.94}\\
				& RF & 90.50 & 90.45 & 87.96 & 90.40 & 91.23 & \textbf{91.86} & & RF & 93.94 & 92.74 & 92.58 & 93.96 & 94.01 & \textbf{94.17}\\
				& SVM & 81.40 & 86.25 & 84.90 & 82.07 & 84.58 & \textbf{88.59} & & SVM & 90.21 & 90.08 & 86.51 & 91.13 & 91.07 & \textbf{91.60}\\
				& XGB & 97.76 & 97.33 & 96.46 & 97.74 & 97.71 & \textbf{97.90} & & XGB & 98.27 & 98.18 & 97.86 & 98.34 & 98.41 & \textbf{98.46}\\
				\hline
				& AB & 76.90 & 79.25 & 73.98 & 79.70 & 79.64 & \textbf{79.90} & & AB & \textbf{85.62} & 85.43 & 83.84 & 85.47 & 85.40 & 85.44\\
				& DT & \textbf{84.07} & 83.92 & 79.68 & 83.65 & 83.95 & 84.05 & & DT & 79.76 & 79.63 & 78.41 & 79.99 & 80.12 & \textbf{80.13}\\
				& ET & 86.31 & 85.69 & 82.07 & 87.03 & 86.87 & \textbf{87.09} & & ET & 84.41 & 84.49 & 82.92 & 85.03 & 84.90 & \textbf{85.06}\\
				& \textit{k}NN & 83.92 & 83.65 & 79.59 & 84.60 & 84.47 & \textbf{84.65} & & \textit{k}NN & 83.91 & \textbf{84.91} & 83.49 & 84.56 & 84.59 & 84.80\\
				phoneme & LR & 66.32 & 66.91 & 64.69 & 67.21 & 67.42 & \textbf{67.47} & wind & LR & 85.12 & 85.18 & 83.90 & 85.28 & 85.28 & \textbf{85.34}\\
				& MLP & 76.56 & 77.23 & 72.94 & 78.25 & 78.27 & \textbf{78.39} & & MLP & \textbf{86.77} & 86.52 & 84.93 & 86.67 & 86.66 & 86.66\\
				& RF & 85.72 & 85.72 & 81.25 & 86.02 & 86.13 & \textbf{86.14} & & RF & 84.66 & 84.65 & 83.39 & 85.10 & 85.11 & \textbf{85.24}\\
				& SVM & 66.88 & 67.21 & 65.20 & 68.15 & 68.47 & \textbf{68.68} & & SVM & 85.09 & 85.12 & 84.24 & 85.32 & 85.30 & \textbf{85.46}\\
				& XGB & \textbf{93.81} & 93.56 & 90.96 & 93.63 & 93.67 & 93.58 & & XGB & 93.72 & 93.71 & 92.51 & 93.69 & \textbf{93.74} & 93.70\\
				\hline
				& AB & 87.16 & \textbf{87.16} & 81.58 & 87.08 & 87.07 & 87.14 & & AB & 73.48 & 74.61 & 73.48 & 74.76 & 74.59 & \textbf{75.18}\\
				& DT & 82.92 & \textbf{83.46} & 78.38 & 83.28 & 83.37 & 83.46 & & DT & 82.37 & 83.46 & 82.55 & 83.76 & 83.55 & \textbf{83.99}\\
				& ET & 83.32 & 84.14 & 79.04 & 85.04 & 85.64 & \textbf{86.25} & & ET & 89.24 & 90.34 & 89.41 & 90.31 & 90.27 & \textbf{90.67}\\
				& \textit{k}NN & 85.72 & 85.51 & 78.95 & 86.08 & 86.41 & \textbf{86.60} & & \textit{k}NN & 86.32 & 89.38 & 86.49 & 89.37 & 89.48 & \textbf{90.34}\\
				ailerons & LR & 87.47 & 87.55 & 80.61 & 87.58 & 87.60 & \textbf{87.64} & eeg-eye & LR & 51.25 & 52.92 & 50.61 & 53.69 & 53.73 & \textbf{53.75}\\
				& MLP & 87.54 & 87.79 & 80.09 & 87.83 & \textbf{87.99} & 87.96 & & MLP & 54.58 & 55.71 & 52.52 & 55.85 & 55.81 & \textbf{56.64}\\
				& RF & 84.46 & 85.60 & 79.29 & 86.05 & 86.39 & \textbf{86.82} & & RF & 88.21 & 89.02 & 87.84 & 89.03 & 88.99 & \textbf{89.34}\\
				& SVM & 87.49 & 87.55 & 80.54 & 87.65 & 87.66 & \textbf{87.75} & & SVM & 54.56 & 56.56 & 54.62 & 56.97 & 56.80 & \textbf{57.23}\\
				& XGB & 95.48 & 95.62 & 90.62 & 95.58 & 95.60 & \textbf{95.64} & & XGB & 91.70 & 93.02 & 91.63 & 92.89 & 92.84 & \textbf{93.28}\\
				\hline
				& AB & 81.01 & 81.89 & 77.72 & 82.53 & 82.53 & \textbf{82.95} & & AB & 93.13 & 93.47 & 92.46 & 93.66 & 93.73 & \textbf{94.04}\\
				& DT & 79.89 & 80.04 & 77.51 & 80.45 & 80.43 & \textbf{80.67} & & DT & 93.26 & 92.90 & 91.95 & 93.06 & 93.14 & \textbf{93.28}\\
				& ET & 82.64 & 81.60 & 80.48 & 83.29 & 83.41 & \textbf{83.76} & & ET & 95.58 & 95.44 & 95.07 & 95.56 & 95.63 & \textbf{95.69}\\
				& \textit{k}NN & 79.55 & 80.17 & 79.40 & 80.70 & 80.89 & \textbf{81.10} & & \textit{k}NN & 94.25 & 94.09 & 93.29 & 94.29 & 94.32 & \textbf{94.44}\\
				magic & LR & 74.24 & 75.84 & 75.84 & 77.00 & 76.92 & \textbf{77.31} & nomao & LR & 93.30 & 93.12 & 91.93 & 93.73 & 93.74 & \textbf{93.76}\\
				& MLP & 83.66 & 83.70 & 80.38 & 84.26 & 84.28 & \textbf{84.43} & & MLP & 95.00 & 94.84 & 91.94 & 94.98 & 95.02 & \textbf{95.19}\\
				& RF & 83.70 & 82.93 & 80.87 & 84.15 & 84.21 & \textbf{84.43} & & RF & 95.53 & 95.33 & 94.27 & 95.49 & 95.52 & \textbf{95.62}\\
				& SVM & 74.09 & 75.92 & 75.34 & 76.94 & 76.89 & \textbf{77.14} & & SVM & 93.54 & 93.42 & 92.01 & 94.01 & 94.03 & \textbf{94.05}\\
				& XGB & 92.14 & 92.21 & 89.74 & 92.68 & 92.70 & \textbf{92.88} & & XGB & 98.95 & 98.97 & 98.73 & 99.01 & 99.05 & \textbf{99.13}\\
				\hline
				& AB & 67.05 & 67.83 & 67.09 & 67.37 & 67.46 & \textbf{68.95} & & AB & 85.81 & 85.86 & 83.88 & 86.52 & 86.48 & \textbf{86.69}\\
				& DT & 70.88 & 70.67 & 68.28 & 70.24 & 70.15 & \textbf{70.94} & & DT & 78.44 & 78.42 & 75.81 & \textbf{79.53} & 79.21 & 79.27\\
				& ET & 64.65 & 67.40 & 65.39 & 68.09 & 68.70 & \textbf{68.77} & & ET & 85.53 & 85.17 & 83.03 & 85.90 & 85.89 & \textbf{86.19}\\
				& \textit{k}NN & 61.49 & 64.75 & \textbf{68.20} & 62.97 & 63.57 & 64.76 & & \textit{k}NN & 78.58 & 81.64 & \textbf{82.19} & 79.24 & 79.97 & 81.88\\
				bank & LR & 65.03 & 65.98 & 61.16 & 66.03 & 66.03 & \textbf{66.28} & vehicle & LR & 85.66 & 85.68 & 83.02 & 85.74 & 85.71 & \textbf{85.96}\\
				& MLP & 72.37 & 72.36 & 68.26 & 71.79 & \textbf{72.70} & 72.26 & & MLP & 86.79 & \textbf{87.10} & 83.27 & 86.63 & 86.48 & 87.03\\
				& RF & 66.54 & 66.29 & 65.51 & 67.53 & 67.71 & \textbf{68.24} & & RF & 85.55 & 85.18 & 82.93 & 86.28 & 86.16 & \textbf{86.50}\\
				& SVM & 63.95 & 64.13 & 60.61 & 64.15 & 64.28 & \textbf{64.44} & & SVM & 85.31 & 85.33 & 82.54 & 85.48 & 85.37 & \textbf{85.81}\\
				& XGB & 91.84 & 91.98 & 90.77 & 91.90 & 92.07 & \textbf{92.29} & & XGB & 91.86 & 92.05 & 90.85 & 92.30 & 92.29 & \textbf{92.47}\\
				\hline
			\end{tabular}
		}
	\end{center}
\end{table*}

\subsection{Discussion}
\label{Discussion}

The time complexity and space complexity of our algorithm is very low and can be adjusted flexibly according to actual needs. Below we will continue to discuss whether the algorithm meets the requirements in section \ref{Introduction}.

\subsubsection{Strong applicability}

Our algorithm is user-friendly and does not require learning a cumbersome model like reinforcement learning and transfer learning based methods. Besides, the hyperparameters which needed to set in advance are only used to control the complexity of the algorithm, such as the number of iterations or iteration time, the number of trees in the forest and the depth of each tree. Therefore, the setting of these hyperparameters is not complicated.

\subsubsection{Distributed computing}

XGBoost is recognized as an algorithm that leverages the parallelism of computing resources, and it has been proven that XGBoost can push the limits of computing power for boosted trees algorithms. At the same time, other aspects of our algorithm can be easily parallelized, such as calculating the information value of the individual feature and the pearson correlation of each feature pair in parallel.

\subsubsection{Real-time inference}

In our algorithm, whether newly generated features can be used for real-time inference depends on the operators $\mathcal{O}$ that used for feature generation. Users can choose different operators according to the actual situation to meet the real-time requirements of the business.

\section{Experiments}
\label{Experiments}

For simplicity and versatility, we only select four basic binary operators $+$, $-$, $\times$ and $\div$ when experimenting with each algorithm.

\begin{table}[H]
	\caption{The information of the benchmark data sets}
	\begin{center}
		\setlength{\tabcolsep}{4mm}{
			\begin{tabular}{lcccc}
				\hline
				Dataset & \#Train & \#Valid & \#Test & \#Dim\\
				\hline
				valley & 900 & - & 312 & 100\\
				banknote & 1,000 & - & 372 & 4\\
				gina & 2,800 & - & 668 & 970\\
				spambase & 3,800 & - & 801 & 57\\
				phoneme & 4,500 & - & 904 & 5\\
				wind & 5,000 & - & 1,574 & 14\\
				ailerons & 9,000 & 2,000 & 2,750 & 40\\
				eeg-eye & 10,000 & 2,000 & 2,980 & 14\\
				magic & 13,000 & 3,000 & 3,020 & 10\\
				nomao & 22,000 & 6,000 & 6,000 & 118\\
				bank & 35,211 & 4,000 & 6,000 & 51\\
				vehicle & 60,000 & 18,528 & 20,000 & 100\\
				\hline
			\end{tabular}
		}
		\label{The information of the public data sets}
	\end{center}
\end{table}

\subsection{Experiments on benchmark data sets}

We first conduct experiments on 12 benchmark data sets, with different sample and feature size. All of these data are available on the OpenML database\footnote{https://www.openml.org/}.  The number of training, validation, and test samples are shown in Table \ref{The information of the public data sets}, with the feature size. Note that for the data set whose sample size is less than 10000, no validation set is splitted, and we simply use training data for validation if necessary.
All experiments are performed on a 4-core computer with 16GB of RAM.

\subsubsection{Algorithms for comparison}

We compare our model with the original features (ORIG), two other state-of-the-art feature generating algorithms, i.e., FCTree \cite{DBLP:conf/sdm/FanZPVZRYY10} and TFC \cite{DBLP:journals/eswa/PiramuthuS09}, and two of our own comparison algorithms, i.e., Random (RAND) and SAFE-Important (IMP). We use Area Under Curve (AUC) as the evaluation metric.

RAND algorithm randomly selects $\gamma$ different feature combinations of all original features for feature generation. Different from it, IMP algorithm only randomly selects $\gamma$ different feature combinations with the split features of XGBoost for feature generation. RAND and IMP follow the same feature selection process as SAFE. For the convenience of comparison, The maximum number of RAND, IMP, and SAFE output features are set to $2M$. Features generated by FCTree will also be reduced to $2M$ according to information gain. Moreover, TFC, RAND, IMP and SAFE only perform one iteration.

\subsubsection{Classification performance}

We evaluate the generated features (and also the original features) of each compared algorithm on 9 state-of-the-art classification algorithms (CLF), which are AdaBoost (AB), Decison Tree (DT), Extremely randomized Trees (ET), $k$ nearest neighbors ($k$NN), Logistic Regression (LR), Multi Layered Perceptron (MLP), Random Forest (RF), SVM with linear kernel (SVM) and XGBoost. All parameters of these algorithms are set as the default values in scikit-learn~\cite{scikit-learn} and XGBoost~\cite{DBLP:conf/kdd/ChenG16}. We performed $n$ times experiments, and obtain the final results by averaging the results of these experiments ($n$ is 100 for the first 9 data sets and 10 for the rest data sets).

The reported performances are measured in terms of AUC, which are shown in Table \ref{Classification performance on benchmark data sets}. The value in the table means $100 \times \text{AUC}$. 
It can be seen from the experimental results that SAFE has a significant advantage over all other compared algorithms no matter what model is performed after the feature generation process. Compared with the original feature space, the features generated by our model can improve the overall prediction AUC by $6.50\%$ on average. Compared with FCTree and TFC, SAFE can improve the performance by $2.03\%$ and $3.74\%$ on average, respectively. What's more, SAFE performs better than RAND and IMP, indicating that our algorithm does mine a combination of features that are more likely to generate better features.

\subsubsection{Feature importance}

\begin{figure*}[htbp]
	\centering
	\subfigure[valley]{
		\begin{minipage}[t]{0.32\linewidth}
			\includegraphics[width=2.2in,trim={0.9cm 0cm 2.0cm 1.1cm},clip]{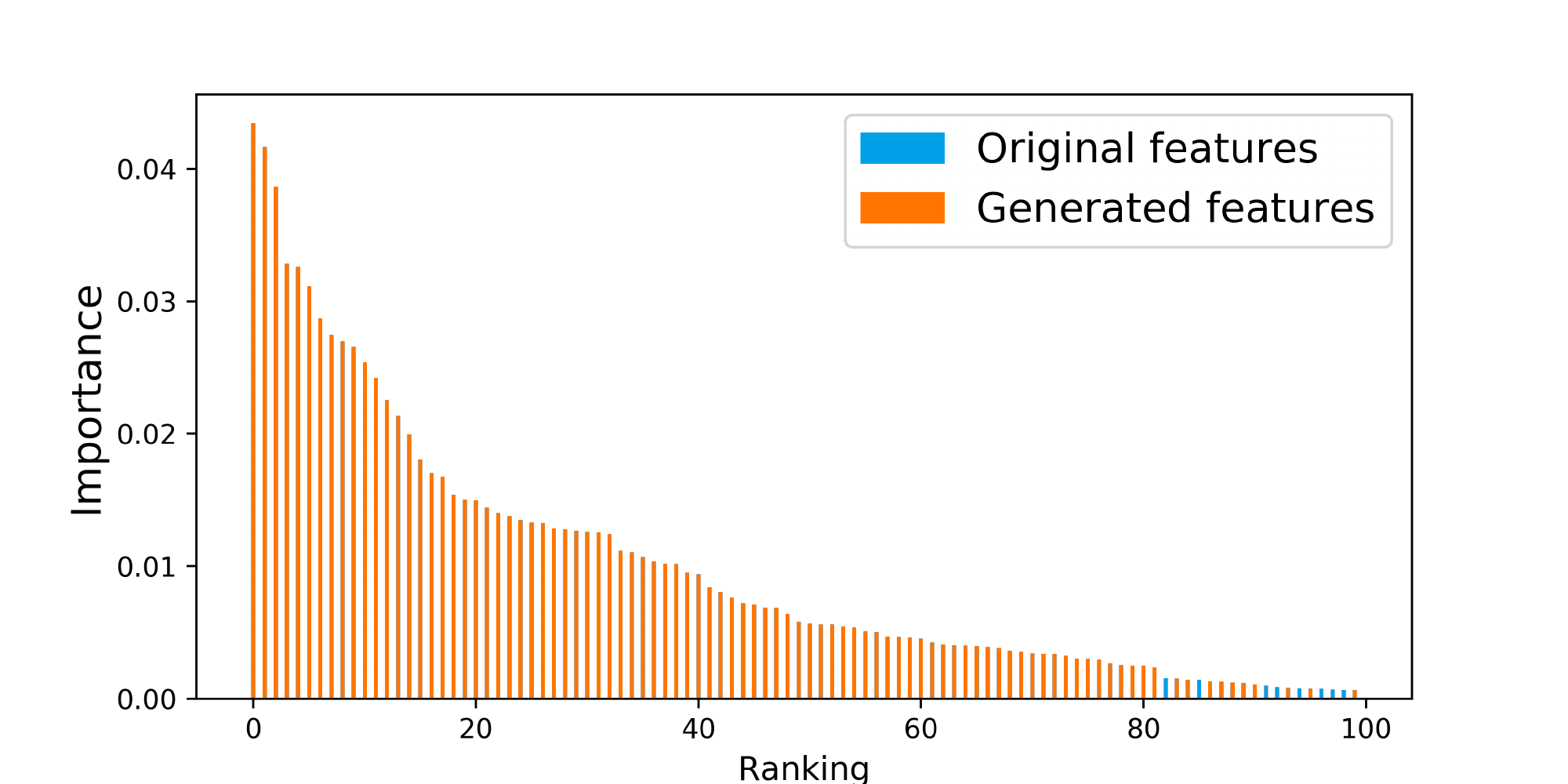}
		\end{minipage}
	}%
	\subfigure[banknote]{
		\begin{minipage}[t]{0.32\linewidth}
			\includegraphics[width=2.2in,trim={0.9cm 0cm 2.0cm 1.1cm},clip]{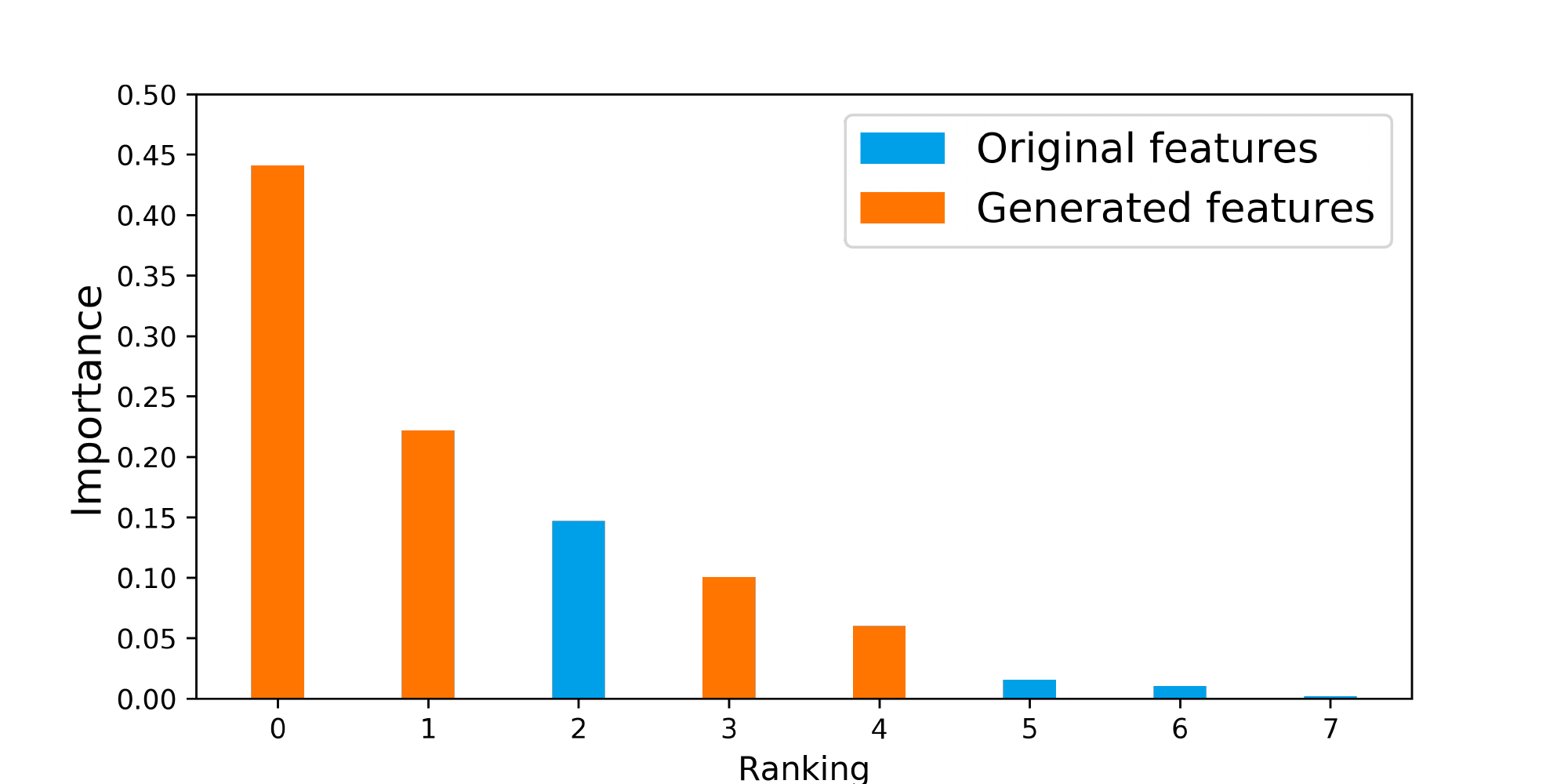}
		\end{minipage}
	}%
	\subfigure[gina]{
		\begin{minipage}[t]{0.32\linewidth}
			\includegraphics[width=2.2in,trim={0.9cm 0cm 2.0cm 1.1cm},clip]{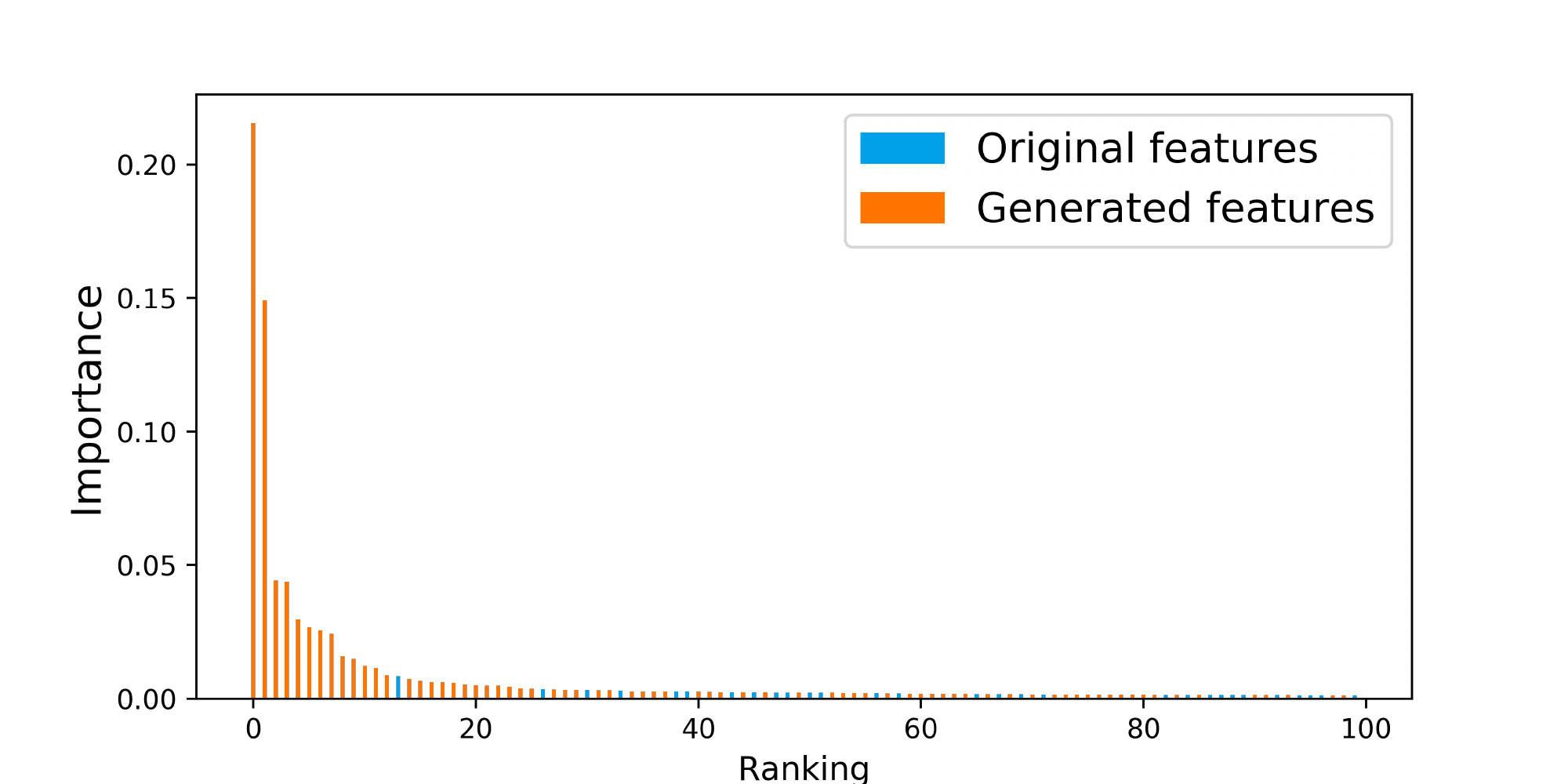}
		\end{minipage}
	}%
	
	\subfigure[spambase]{
		\begin{minipage}[t]{0.32\linewidth}
			\includegraphics[width=2.2in,trim={0.9cm 0cm 2.0cm 1.1cm},clip]{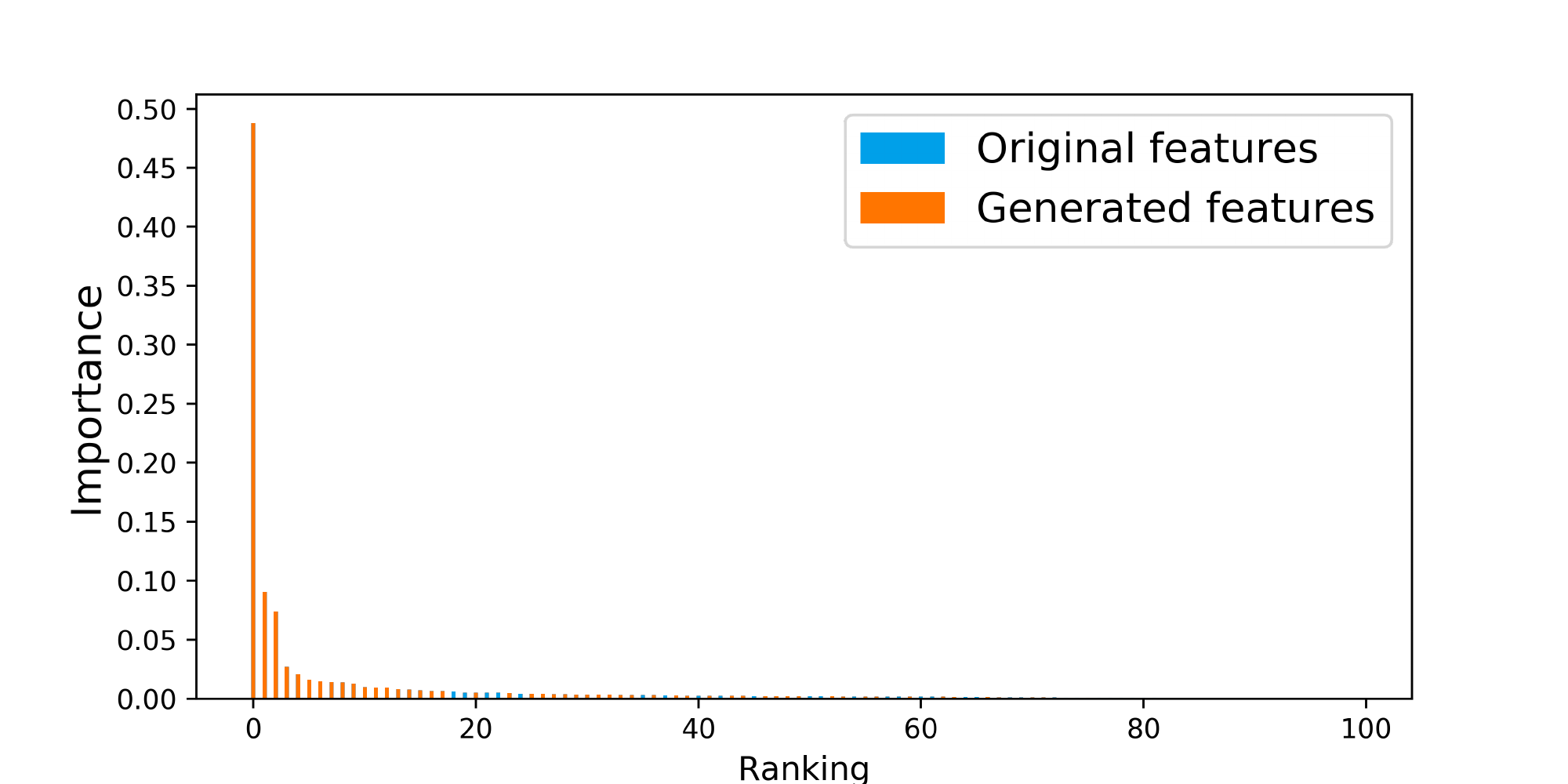}
		\end{minipage}
	}%
	\subfigure[phoneme]{
		\begin{minipage}[t]{0.32\linewidth}
			\includegraphics[width=2.2in,trim={0.9cm 0cm 2.0cm 1.1cm},clip]{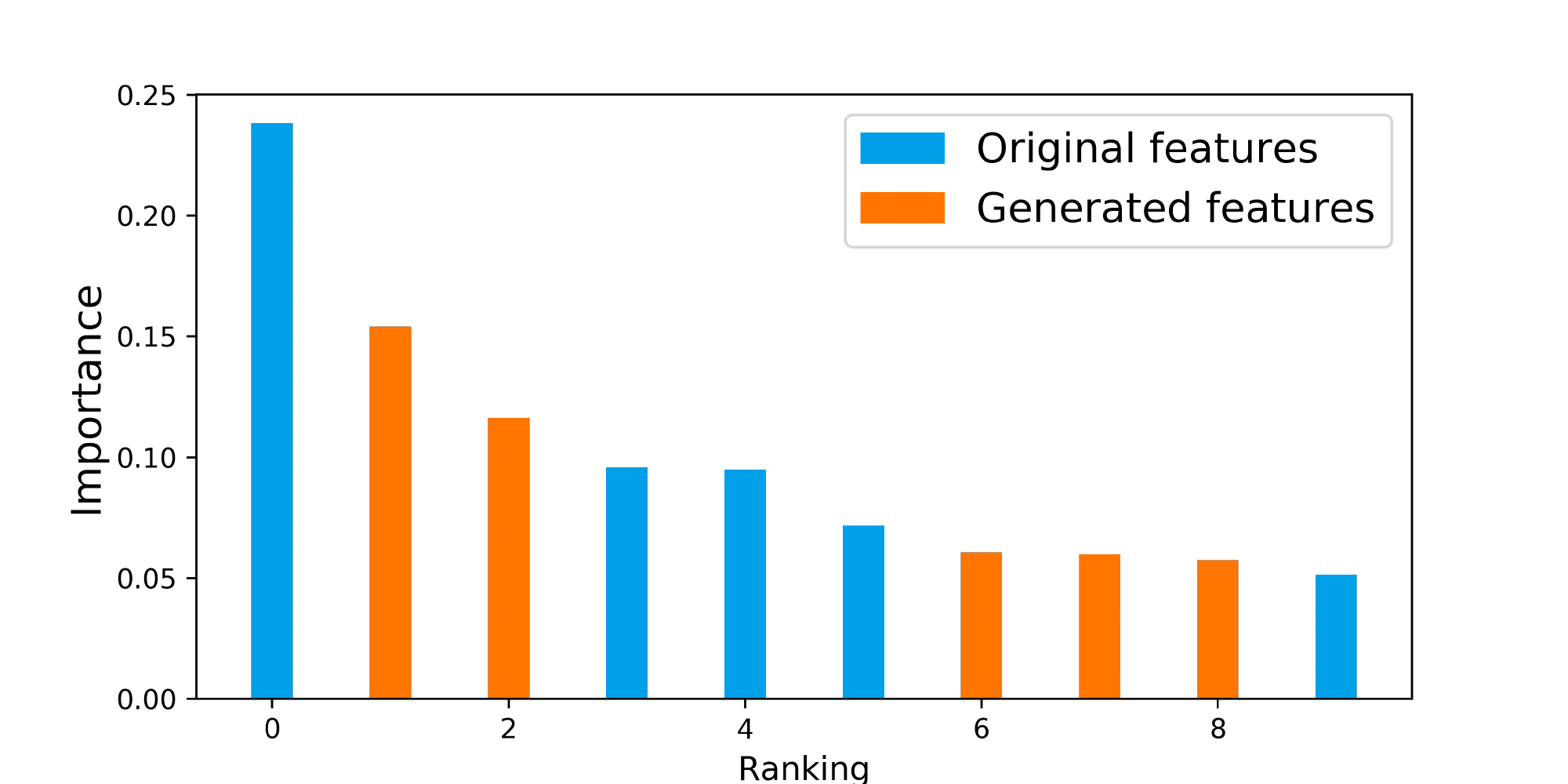}
		\end{minipage}
	}%
	\subfigure[wind]{
		\begin{minipage}[t]{0.32\linewidth}
			\includegraphics[width=2.2in,trim={0.9cm 0cm 2.0cm 1.1cm},clip]{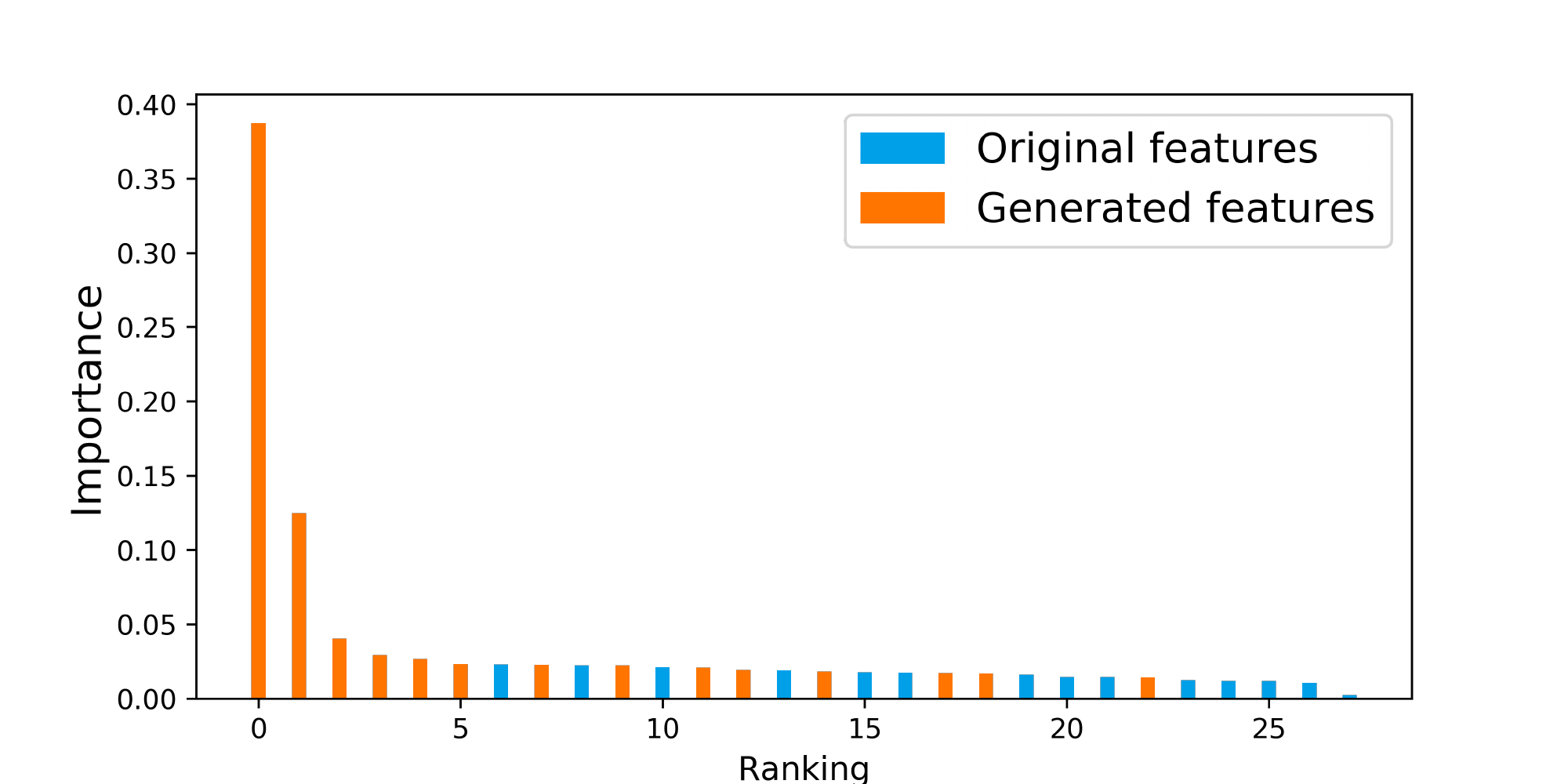}
		\end{minipage}
	}%
	
	\subfigure[ailerons]{
		\begin{minipage}[t]{0.32\linewidth}
			\includegraphics[width=2.2in,trim={0.9cm 0cm 2.0cm 1.1cm},clip]{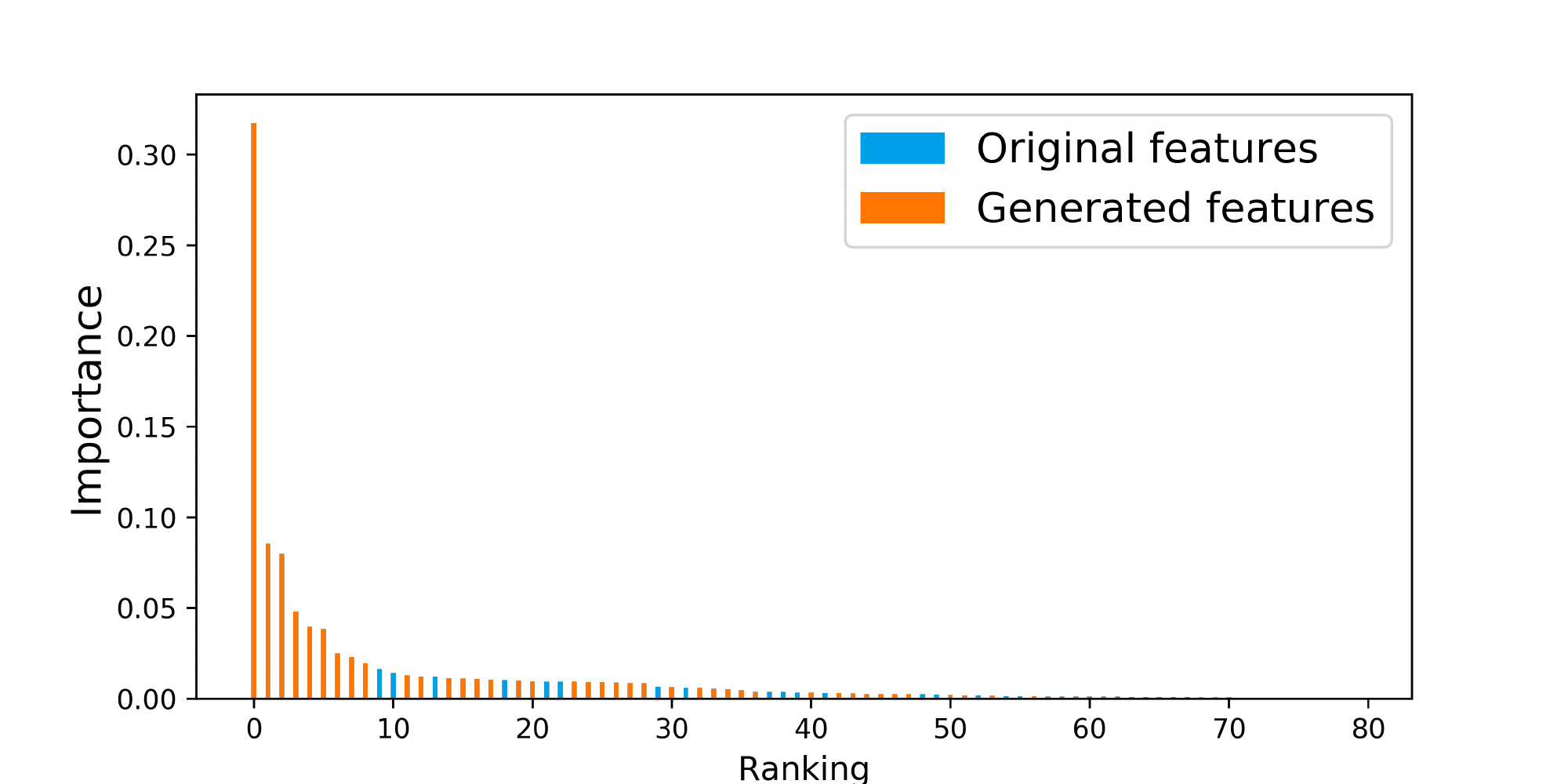}
		\end{minipage}
	}%
	\subfigure[eeg-eye]{
		\begin{minipage}[t]{0.32\linewidth}
			\includegraphics[width=2.2in,trim={0.9cm 0cm 2.0cm 1.1cm},clip]{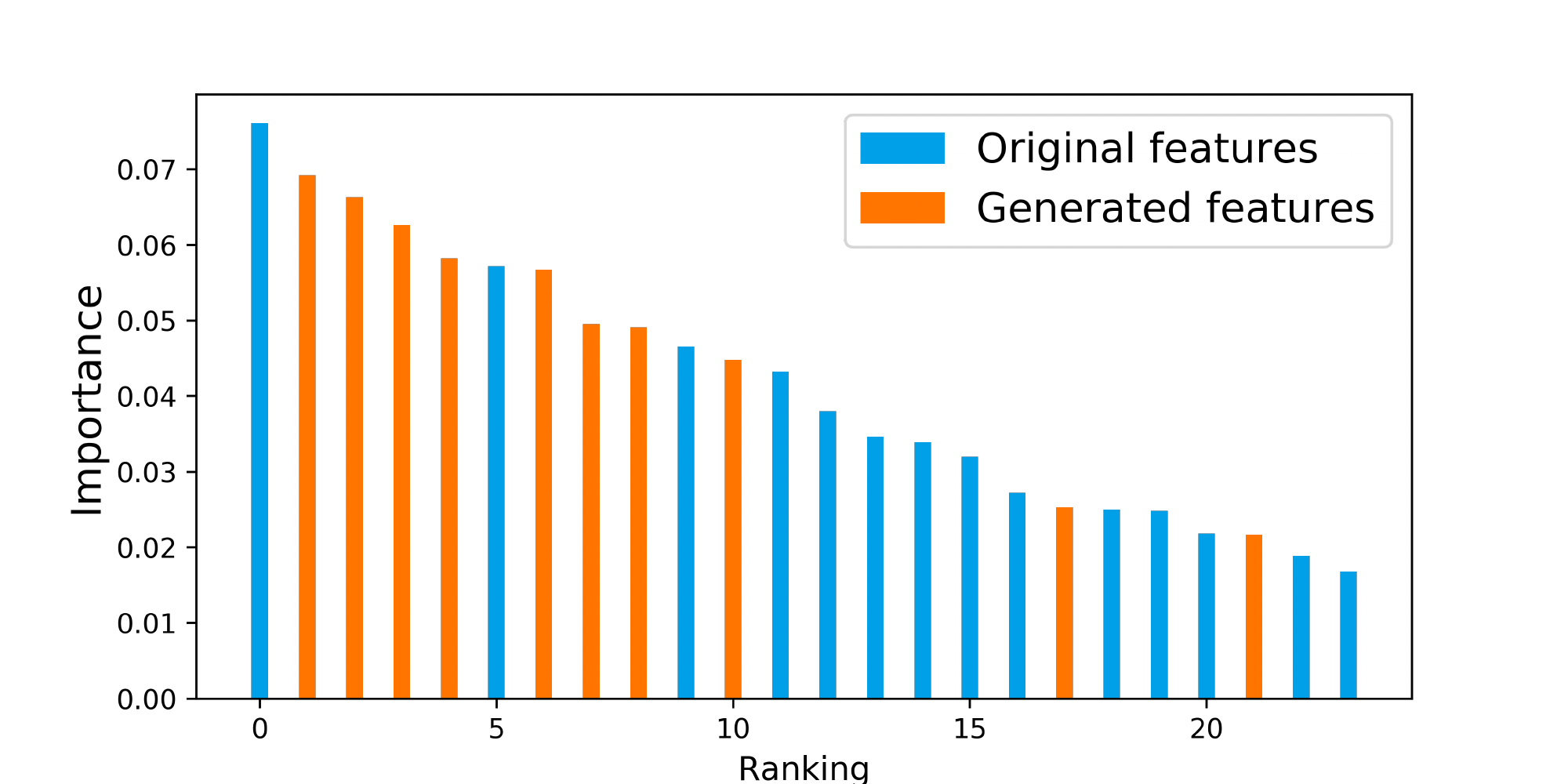}
		\end{minipage}
	}%
	\subfigure[magic]{
		\begin{minipage}[t]{0.32\linewidth}
			\includegraphics[width=2.2in,trim={0.9cm 0cm 2.0cm 1.1cm},clip]{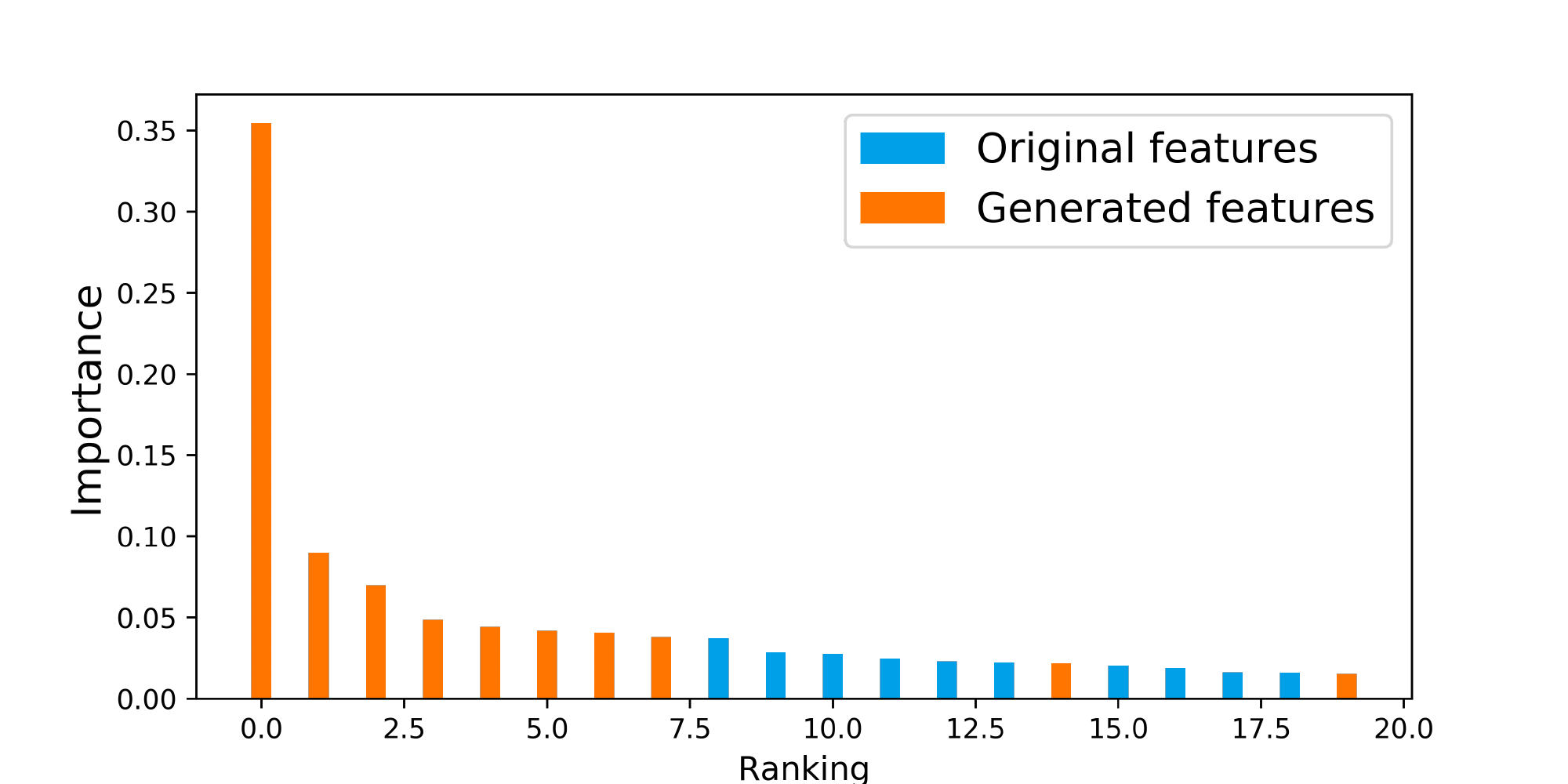}
		\end{minipage}
	}%
	
	\subfigure[nomao]{
		\begin{minipage}[t]{0.32\linewidth}
			\includegraphics[width=2.2in,trim={0.9cm 0cm 2.0cm 1.1cm},clip]{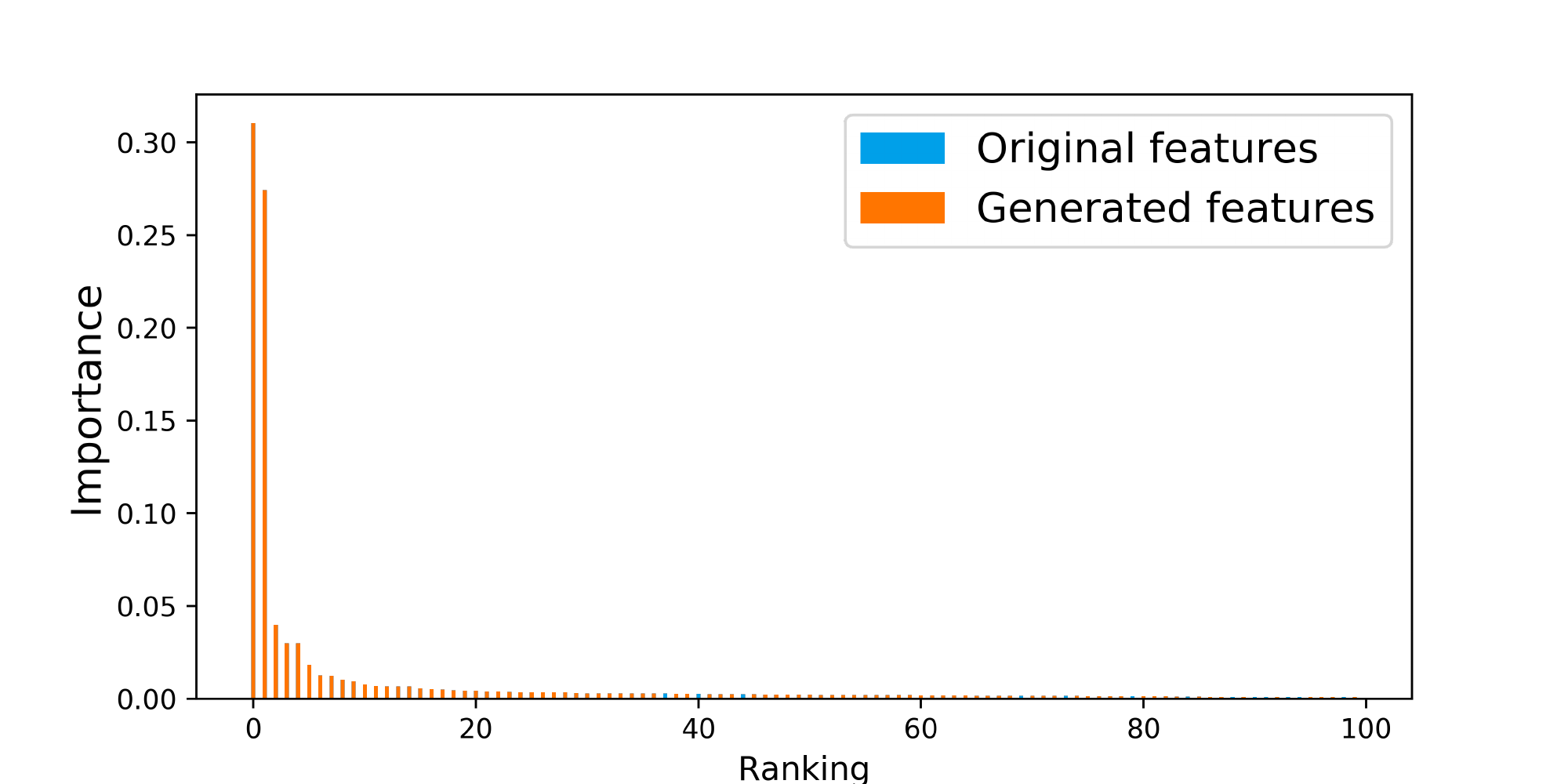}
		\end{minipage}
	}%
	\subfigure[bank]{
		\begin{minipage}[t]{0.32\linewidth}
			\includegraphics[width=2.2in,trim={0.9cm 0cm 2.0cm 1.1cm},clip]{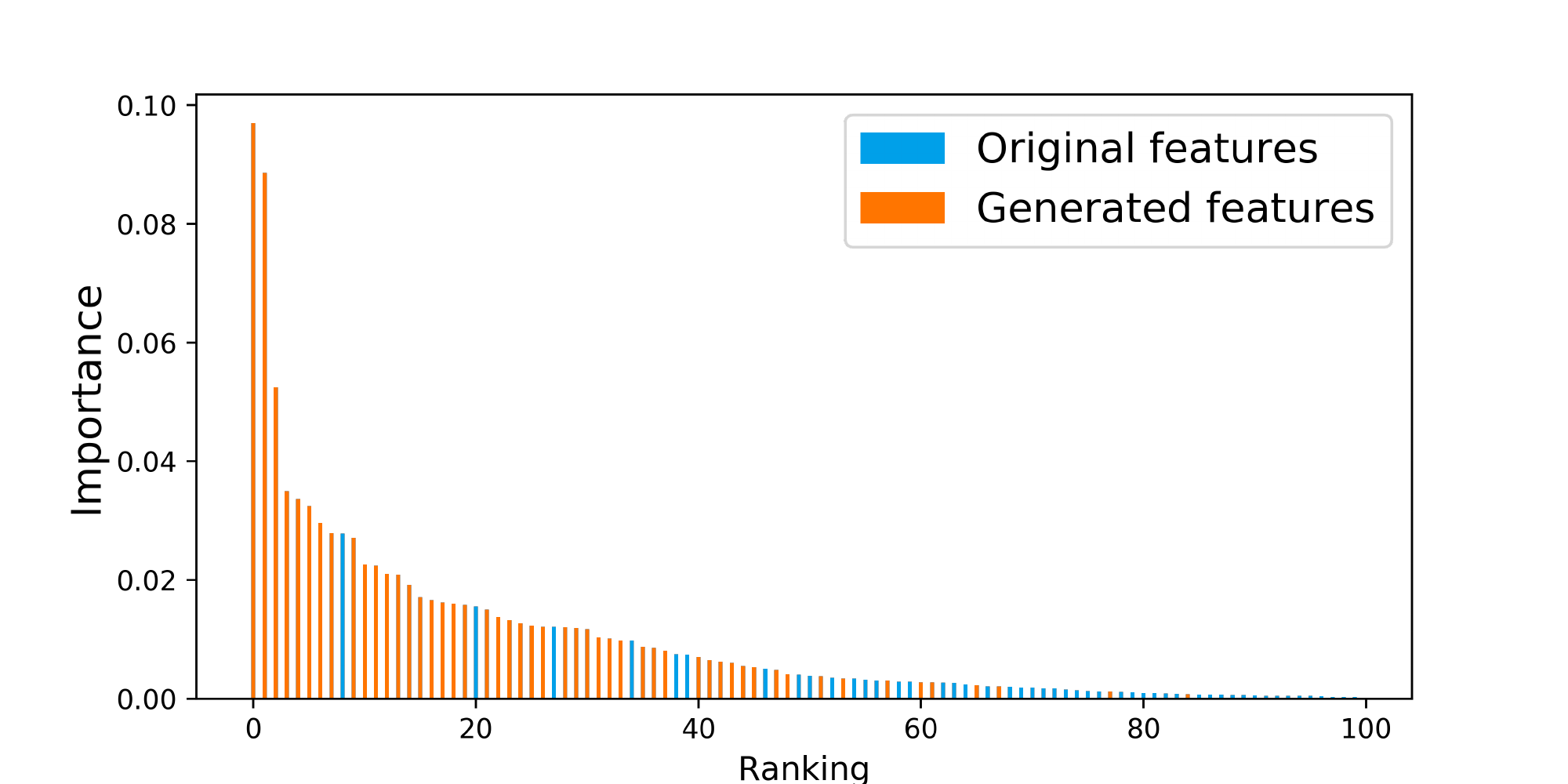}
		\end{minipage}
	}%
	\subfigure[vehicle]{
		\begin{minipage}[t]{0.32\linewidth}
			\includegraphics[width=2.2in,trim={0.9cm 0cm 2.0cm 1.1cm},clip]{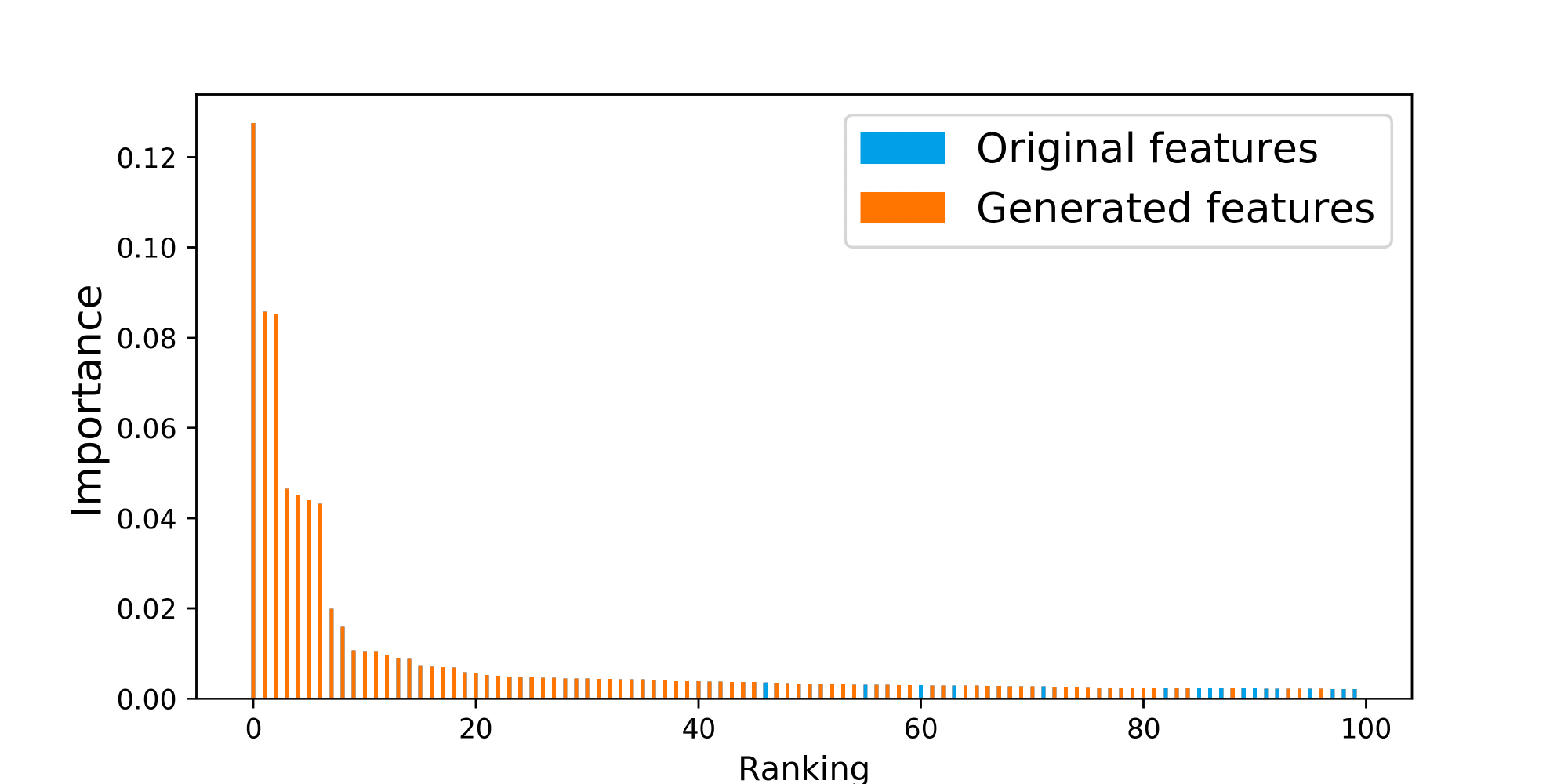}
		\end{minipage}
	}%
	\caption{Feature importance}
	\label{Feature importance}
\end{figure*}

We compare the importance of generated features with original features. We combine the $M$ original features with the top-ranked generated features (up to $M$) to form a new data set and use random forest to score feature importance. The experimental results are shown in Fig. \ref{Feature importance}. It is evident that the new features generated by SAFE (indicated in orange) are relatively more important than the original features (indicated in blue), which validates the effectiveness of the generated features.

\subsubsection{Execution time}

Section \ref{Time complexity} has analyzed the time complexity of each method. Table \ref{Executing time} lists the actual executing time. It can be seen that SAFE has a great advantage and the execution time of it is on average $0.13$ ($0.08$) times the execution time of FCTree (TFC).

\begin{table}[htbp]
	\caption{Execution time (in second)}
	\begin{center}
		\setlength{\tabcolsep}{2.5mm}{
			\begin{tabular}{lccccc}
				\hline
				Dataset & FCT & TFC & RAND & IMP & SAFE\\
				\hline
				valley & 9.80 & 228.93 & 0.55 & 0.65 & 0.73\\
				banknote & 0.16 & 0.27 & 0.08 & 0.31 & 0.12\\
				gina & 84.03 & 95.73 & 3.39 & 5.28 & 5.31\\
				spambase & 23.84 & 262.11 & 2.85 & 3.23 & 3.17\\
				phoneme & 10.83 & 2.48 & 0.46 & 0.58 & 0.51\\
				wind & 25.03 & 22.87 & 2.13 & 2.39 & 2.33\\
				ailerons & 80.73 & 336.53 & 2.12 & 2.57 & 2.72\\
				egg-eye & 58.88 & 42.13 & 1.09 & 1.20 & 1.18\\
				magic & 52.45 & 36.79 & 2.55 & 2.96 & 3.32\\
				nomao & 104.59 & 1469.19 & 22.22 & 26.61 & 28.82\\
				bank & 838.17 & 552.48 & 13.70 & 13.81 & 12.28\\
				vehicle & 1355.19 & 2748.89 & 52.86 & 40.34 & 62.14\\
				\hline
			\end{tabular}
		}
		\label{Executing time}
	\end{center}
\end{table}

\subsubsection{Feature stability}
We further compare the stability of the generated features of each algorithm. The basic idea is that the generated features are more stable if the same features are generated each time when we repeat the automatic feature engineering procedure; and if the features generated each time are different, then the stability of the generated features is unsatisfactory.

Suppose we have conducted $T$ experiments, each time the automatic feature engineering algorithm will generate $2M$ features. Therefore, a total of $2MT$ features will be generated. Their distribution can be expressed as $Dis=\left\{(x^{(s_1)}, t_1),\cdots,(x^{(s_k)}, t_k)\right\}$, where $s_i$ represents the feature id, $t_i$ represents the number of occurrences of the feature and $t_1 \geq t_2 \geq \cdots \geq t_k$. Therefore, the distribution with the best feature stability is $\hat{Dis}=\left\{(x^{(s_1)}, T),\cdots,(x^{(s_{2M})}, T)\right\}$, and the worst is $\widetilde{Dis}=\left\{(x^{(s_1)}, 1),\cdots,(x^{(s_{2MT})}, 1)\right\}$.

We use Jensen-Shannon Divergence (JSD)~\cite{DBLP:journals/tit/Lin91} to evaluate the stability of the feature distribution generated by different automatic feature engineering algorithms. JSD is a variant of Kullback-Leibler divergence (KLD)~\cite{DBLP:books/daglib/0013517}, which is converted as:
\begin{equation}
\centering
JSD(P||Q)=\frac{1}{2} \times (KLD(P||R)+KLD(Q||R))
\end{equation}
where $R=\frac{1}{2} \times (P+Q)$ and KLD is calculated as:
\begin{equation}
\centering
KLD(P||Q)=\sum\limits_iP(i)ln\frac{P(i)}{Q(i)}
\end{equation}

\begin{figure*}[htbp]
	\centering
	\subfigure[valley]{
		\begin{minipage}[t]{0.32\linewidth}
			\centering
			\includegraphics[width=2.2in,trim={1.0cm 0cm 2.0cm 1.1cm},clip]{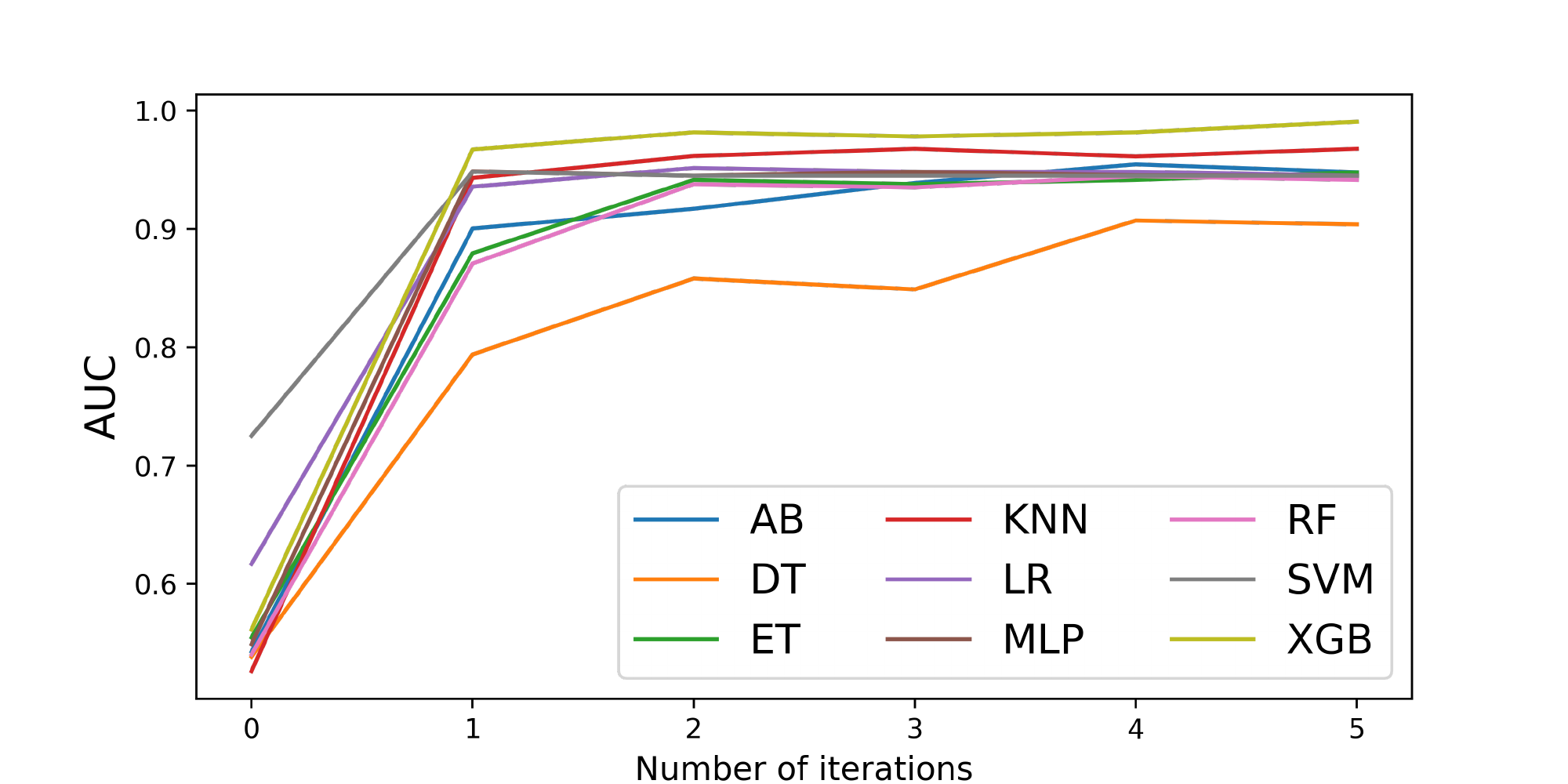}
		\end{minipage}%
	}%
	\subfigure[banknote]{
		\begin{minipage}[t]{0.32\linewidth}
			\centering
			\includegraphics[width=2.2in,trim={0.8cm 0cm 2.0cm 1.1cm},clip]{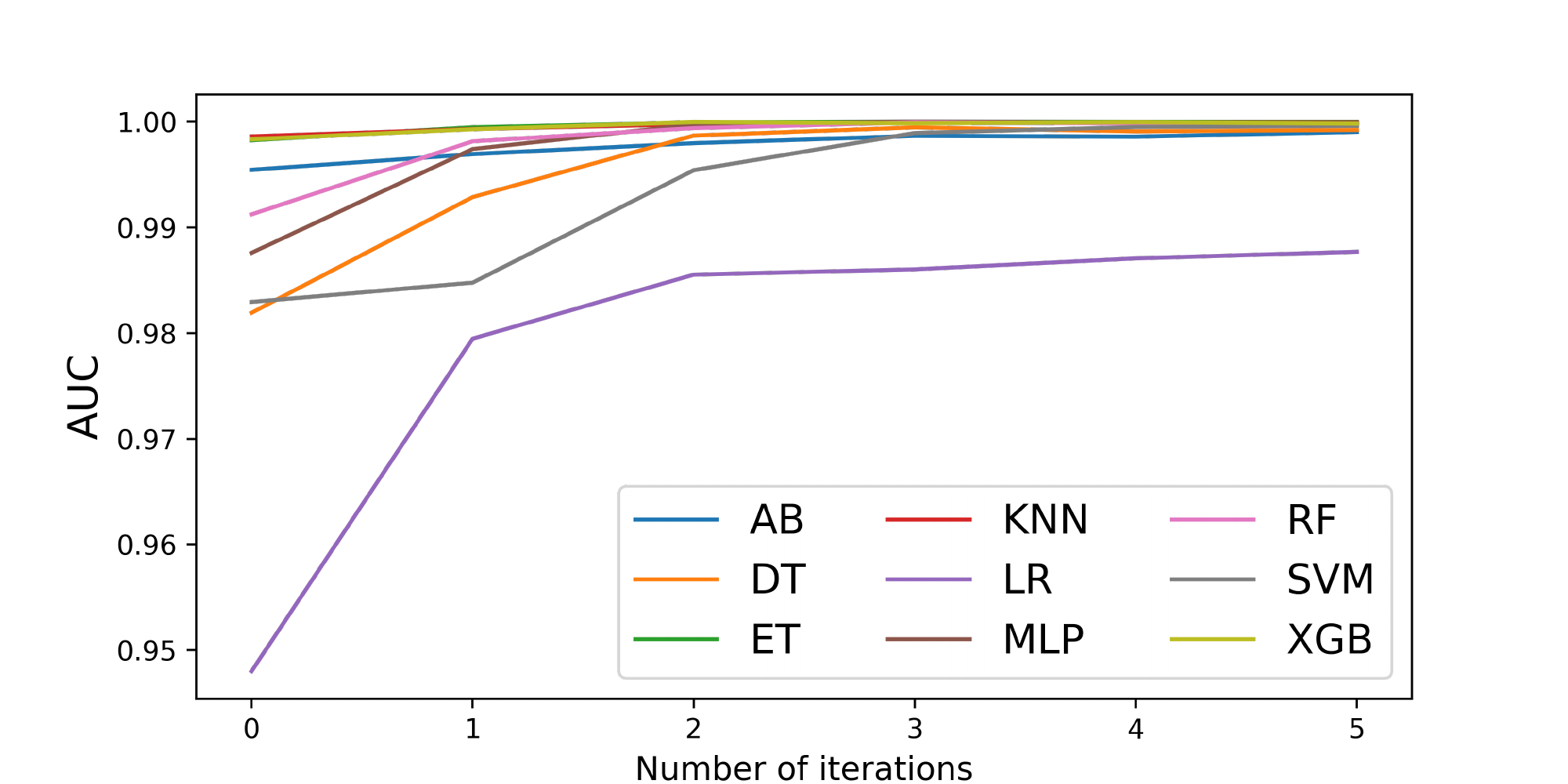}
		\end{minipage}%
	}%
	\subfigure[gina]{
		\begin{minipage}[t]{0.32\linewidth}
			\centering
			\includegraphics[width=2.2in,trim={0.55cm 0cm 2.0cm 1.1cm},clip]{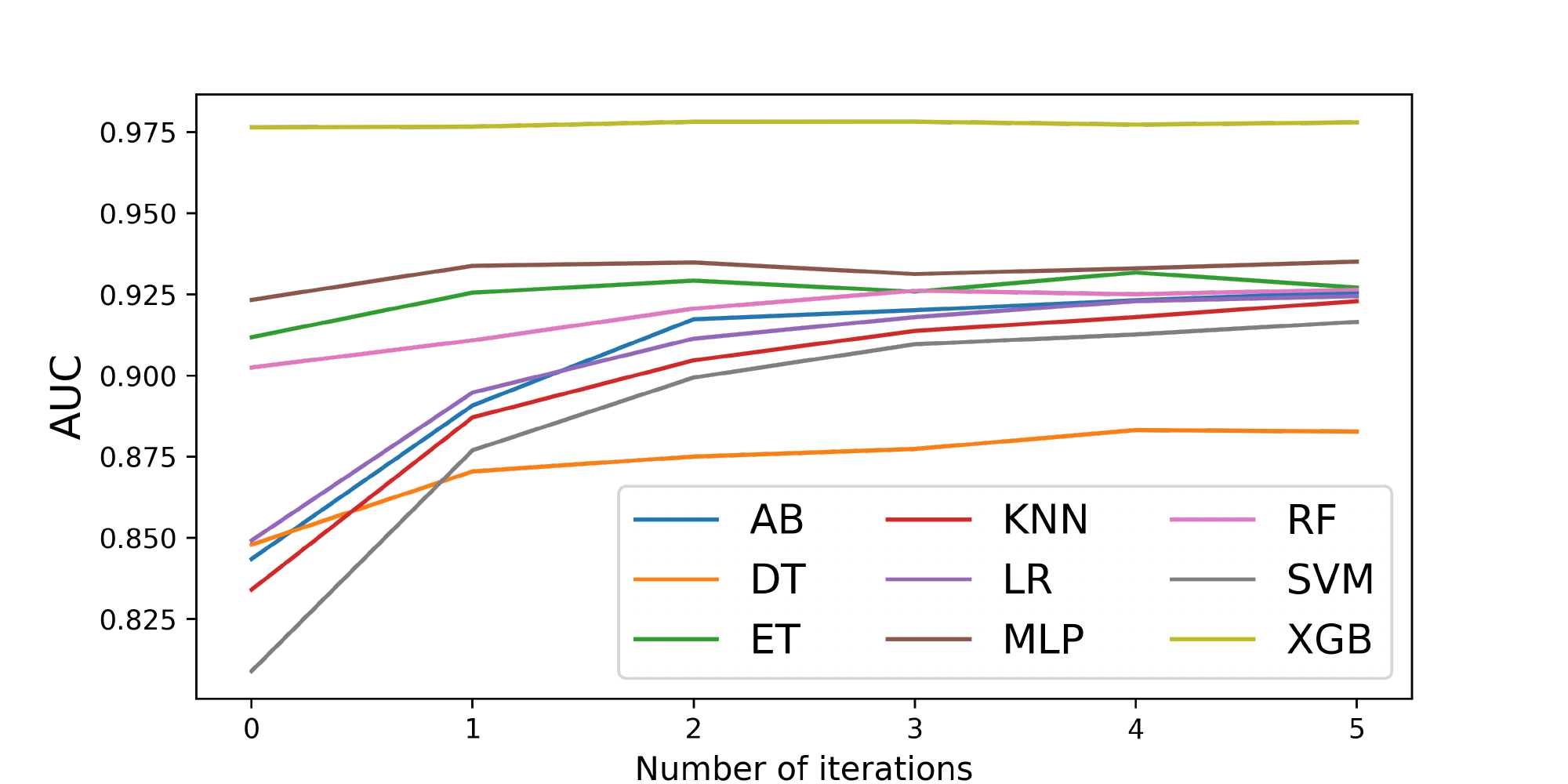}
		\end{minipage}%
	}%
	\centering
	\caption{Performance at different iterations}
	\label{Performance at different iterations}
\end{figure*}

We take $T$ as $100$ and calculate the stability of the generated features of each algorithm. That is, the JSD between the actual feature distribution $Dis$ and the ideal distribution $\hat{Dis}$. The smaller the value is, the better it is. The experimental results are shown in Table \ref{Feature stability}. We did not compare the TFC algorithm because the execution time of TFC is too long, so it is difficult to calculate so many times. From the experimental results, it can be seen that the stability of the generated features of SAFE has certain advantages.

\begin{table}[htbp]
	\caption{Feature stability}
	\begin{center}
		\setlength{\tabcolsep}{4mm}{
			\begin{tabular}{lcccc}
				\hline
				Dataset & FCT & RAND & IMP & SAFE\\
				\hline
				valley & 0.6991 & 0.4933 & 0.4947 & \textbf{0.4710}\\
				banknote & 0.4405 & 0.3233 & 0.3174 & \textbf{0.1197}\\
				gina & 0.5700 & 0.4639 & 0.4739 & \textbf{0.4163}\\
				spambase & 0.5101 & 0.4571 & 0.4399 & \textbf{0.3587}\\
				phoneme & \textbf{0.1947} & 0.3269 & 0.3294 & 0.2616\\
				wind & 0.3707 & 0.3608 & 0.3570 & \textbf{0.3230}\\
				ailerons & 0.4440 & 0.4590 & 0.3963 & \textbf{0.3330}\\
				egg-eye & 0.3529 & 0.3768 & 0.3741 & \textbf{0.3212}\\
				magic & \textbf{0.1847} & 0.3306 & 0.3384 & 0.2620\\
				nomao & 0.5061 & 0.5032 & 0.4735 & \textbf{0.4065}\\
				bank & 0.3713 & 0.4240 & 0.4072 & \textbf{0.2853}\\
				\hline
			\end{tabular}
		}
		\label{Feature stability}
	\end{center}
\end{table}

\subsubsection{Performance at different iterations}

We then validate whether the performance can be further improved as the iteration process goes on. We set the iteration round to 5, and the sampled results are shown in Fig.~\ref{Performance at different iterations}. As we can see, the performance may further be improved as the round proceeds, and become stable after some rounds. This is reasonable, since that in the first some rounds, more useful feature combinations can be excavated so that the performance can be further improved, and after some rounds, there may be no new useful feature combinations can be found, thus the features will not be updated, and the performance keeps unchanged.

\subsection{Experiments on business data sets}

Experiments on extra-large scale business data sets are further conducted to verify the effectiveness and scalability of the proposed method on real industrial tasks. 
The data sets come from the tasks for fraud detection in Ant Financial, which aims at finding the potential fraud transactions (or malicious users), so that the system can stop these transactions (or catch these users) to avoid the economic losses.
Table \ref{table_info_business_data} presents the detailed information of these data sets. As we can see, the number of samples is extremely large (e.g., up to 8 million training samples for Data3), which severely hinders the employment of many preceding state-of-the-art methods.
All parameters of the evaluated models are set as the default values as before.

\begin{table}[htbp]
	\caption{The information of the bussiness data sets}
	\label{table_info_business_data}
	\begin{center}
		\setlength{\tabcolsep}{3mm}{
			\begin{tabular}{lcccc}
				\hline
				Dataset & \#Train & \#Valid & \#Test & \#Dim\\
				\hline
				Data1 & 2,502,617 & 625,655 & 625,655 & 81\\
				Data2 & 7,282,428 & 1,820,607 & 1,820,607 & 44\\
				Data3 & 8,000,000 & 2,000,000 & 2,000,000 & 73\\
				
				\hline
			\end{tabular}
		}
	\end{center}
\end{table}

The results are shown in table~\ref{table_performance_for_business_data}. TFC and FCTree are not compared since the execution time is too long for these two methods when applying for these extremely large scale data sets. 
Thanks to the delicate design in the feature generation procedure of SAFE, the whole time consuming is acceptable even for the industrial tasks.
More important, as we can see, the proposed method SAFE can consistently improve the performance, which validate the effectiveness of the proposed method when applying in real industrial tasks and make it a choice for performing automatic feature engineering for extremely large scale industrial data sets. Actually, this framework has been deployed in our system, providing help for many different real-world tasks.

\begin{table}[htbp]
	\caption{Classification performance of business data sets}
	\label{table_performance_for_business_data}
	\begin{center}
		\setlength{\tabcolsep}{3mm}{
			\begin{tabular}{cccccc}
				\hline
				Dataset & CLF & ORIG & RAND & IMP & SAFE\\
				\hline
				
				& LR & 93.07 & 95.81 & 95.83 & \textbf{95.93}\\
				Data1 & RF & 96.26 & 97.62 & 97.60 & \textbf{98.20}\\
				& XGB & 97.04 & 96.59 & 97.35 & \textbf{97.46}\\
				\hline
				& LR & 90.24 & 90.26 & 90.26 & \textbf{90.31}\\
				Data2 & RF & 88.26 & 88.71 & 88.61 & \textbf{88.95}\\
				& XGB & 90.13 & 90.33 & 90.44 & \textbf{90.61}\\
				\hline
				& LR & 89.64 & 89.82 & 89.84 & \textbf{89.94}\\
				Data3 & RF & 86.98 & 87.05 & 88.26 & \textbf{88.59}\\
				& XGB & 89.77 & 89.74 & 89.92 & \textbf{90.37}\\
				\hline
			\end{tabular}
		}
	\end{center}
\end{table}

\section{Conclusion}
\label{Conclusion}
Automatic feature engineering has become an important topic of autoML in recent years, and many different methods have been proposed to handle this task.
However, the efficiency and scalability of these methods are still far from satisfactory, especially for industrial tasks, while automatically performing feature engineering is severely demanded. 
In this paper, we propose a scalable and efficient method named SAFE for automatic feature engineering.
Extensive experiments on both benchmark data sets and extra-large scale business data sets are conducted, and detailed analysis is provided, which shows that the proposed method can provide prominent efficiency and competitive effectiveness when comparing with other methods.

\bibliographystyle{IEEEtran}
\bibliography{autoFeature-ref-new} 

\end{document}